\documentclass[runningheads]{llncs}
\usepackage{graphicx}
\usepackage{tikz}
\usepackage{comment}
\usepackage{amsmath,amssymb} % define this before the line numbering.
\usepackage{color}
\usepackage{bm}
\usepackage{multirow}
\usepackage{subfigure,cases}
\usepackage{booktabs}
\usepackage{url}
\usepackage{xcolor}
\usepackage{orcidlink}
\usepackage[ruled,vlined]{algorithm2e}

\definecolor{commentcolor}{RGB}{110,154,155}   % define comment color
\newcommand{\PyComment}[1]{\ttfamily\textcolor{commentcolor}{\# #1}}  % add a "#" before the input text "#1"
\newcommand{\PyCode}[1]{\ttfamily\textcolor{black}{#1}} % \ttfamily is the code font

\usepackage[accsupp]{axessibility}  % Improves PDF readability for those with disabilities.

\usepackage{hyperref}
\hypersetup{hidelinks}

\begin{document}
% \linenumbers
\pagestyle{headings}
\mainmatter
\def\ECCVSubNumber{3080}  % Insert your submission number here

\title{Doubly Deformable Aggregation of Covariance Matrices for Few-shot Segmentation} % Replace with your title

%\begin{comment}
% \titlerunning{ECCV-22 submission ID \ECCVSubNumber} 
% \authorrunning{ECCV-22 submission ID \ECCVSubNumber} 
% \author{Anonymous ECCV submission}
% \institute{Paper ID \ECCVSubNumber}
%\end{comment}
%******************

% \begin{comment}
\titlerunning{DACM for few-shot segmentation}
\author{Zhitong Xiong \orcidlink{0000-0002-3953-585X} \inst{1} \and
Haopeng Li \orcidlink{0000-0001-8175-5381} \inst{2} \and
Xiao Xiang Zhu \orcidlink{0000-0001-5530-3613} \inst{1,3}}
\authorrunning{Z. Xiong et al.}
\institute{Data Science in Earth Observation, Technical University of Munich (TUM)\\ \and
School of Computing and Information Systems, University of Melbourne \and
Remote Sensing Technology Institute (IMF), German Aerospace Center (DLR)\\}
% \end{comment}
%******************
\maketitle

\begin{abstract}
Training semantic segmentation models with few annotated samples has great potential in various real-world applications. For the few-shot segmentation task, the main challenge is how to accurately measure the semantic correspondence between the support and query samples with limited training data. To address this problem, we propose to aggregate the learnable covariance matrices with a deformable 4D Transformer to effectively predict the segmentation map. Specifically, in this work, we first devise a novel hard example mining mechanism to learn covariance kernels for the Gaussian process. The learned covariance kernel functions have great advantages over existing cosine similarity-based methods in correspondence measurement. Based on the learned covariance kernels, an efficient doubly deformable 4D Transformer module is designed to adaptively aggregate feature similarity maps into segmentation results. By combining these two designs, the proposed method can not only set new state-of-the-art performance on public benchmarks, but also converge extremely faster than existing methods. Experiments on three public datasets have demonstrated the effectiveness of our method.\footnote{Code: \url{https://github.com/ShadowXZT/DACM-Few-shot.pytorch}}

\end{abstract}

\section{Introduction}
Semantic segmentation at the pixel level \cite{zheng2021rethinking,yang2020fda,long2015fully,hao2020brief} is one of the fundamental tasks in computer vision and has been extensively studied for decades. In recent years, the performance of semantic segmentation tasks has been significantly improved due to the substantial progress in deep learning techniques. Large-scale convolutional neural networks (CNNs) \cite{simonyan2014very,he2016deep}, vision Transformers \cite{dosovitskiy2020image,liu2021swin}, and MLP-based deep networks \cite{liu2021pay} have greatly improved the ability of visual representation learning, and significantly enhanced the performance of downstream tasks such as semantic segmentation \cite{zhu2021unified,zheng2021rethinking}. %The other reason is the availability of large-scale datasets with ground truth semantic segmentation annotations. Segmentation datasets with a large amount of training data make it possible to train modern deep neural networks effectively. 

\begin{figure}[t]
	\centering
	\includegraphics[width=0.831\textwidth]{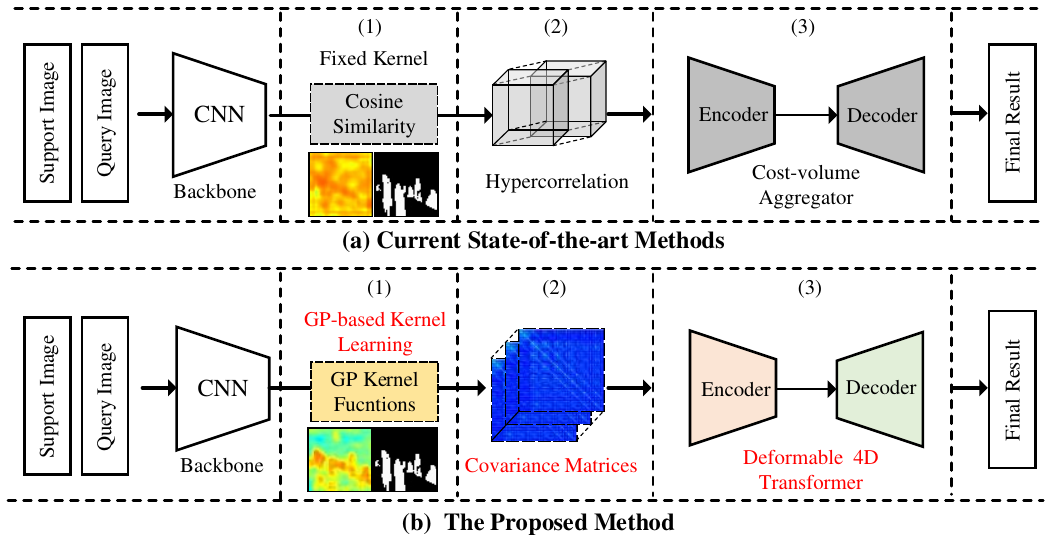}
	\caption{Comparison of the proposed DACM (b) with the framework of existing state-of-the-art methods (a). There are three main differences. (1) We propose to utilize GP-based kernel functions for similarity measurement. (2) We use covariance matrices instead of cosine similarity as the cost volume. (3) A DDT module is designed for effective cost volume aggregation.
}
	\label{compare}
\end{figure} 

However, these advances currently rely heavily on large-scale annotated datasets that require extensive manual annotation efforts. This is not conducive to real-world computer vision applications, as obtaining pixel-level annotations is expensive and time-consuming. To remedy this issue, the few-shot segmentation task has emerged and attracted more and more research attention \cite{yang2020prototype,wang2019panet,liu2020part,yang2020new,hu2019attention,zhang2019pyramid}. For few-shot segmentation, only a handful of annotated samples (support samples) for each class are provided for the training of the model, which largely mitigates the reliance on large-scale manual annotations. Although an attempt has been made \cite{boudiaf2021few} to handle few-shot segmentation in a transductive way, most existing works model the few-shot segmentation in a meta-learning setting \cite{shaban2017one,vinyals2016matching} to avoid overfitting 
brought by insufficient training data. %In this setting, the training and testing subsets consist of non-overlapping classes (tasks). the segmentation model is required to learn across different classes of objects and generalize to unseen classes. 
As support samples are the most important guidance for the prediction of query samples, the key to few-shot segmentation is to effectively exploit information from the support set.

Prototype-based few-shot segmentation methods attempt to extract a prototype vector to represent the support sample \cite{wang2021variational,sun2021attentional,yang2020prototype}.
Despite its effectiveness, compressing the features of support samples into a single vector is prone to noise and neglects the spatial structure of objects in the support image. To maintain the geometric structure of objects, visual correspondence-based methods \cite{min2021hypercorrelation,hu2019attention,zhang2019pyramid} opt to utilize dense, pixel-wise semantic correlations between support and query images. Sophisticated pixel-wise attention-based methods \cite{zhang2021few} have also been devised to leverage dense correlations between support and query samples. However, directly using pixel-level dense information is a double-edged sword, which brings two new challenges: feature ambiguities caused by repetitive patterns, and background noise. To handle these issues, recent work \cite{min2021hypercorrelation} shows that aggregating dense correlation scores (also called \emph{cost volume}) with 4D convolution can attain outstanding performance for few-shot segmentation. To better model the interaction among correlation scores, \cite{hong2021cost} proposes to combine 4D convolutions and 4D swin Transformer \cite{liu2021swin} in a coarse-to-fine manner. Although recent cost aggregation-based methods \cite{min2021hypercorrelation,hong2021cost} attain state-of-the-art performance, they still suffer from two limitations: 1) \emph{lack of flexibility in computing correlation scores}, which results in an extremely slow convergence speed; 2) \emph{lack of flexibility in aggregating high-dimensional correlation scores}, which leads to limited performance with high computational cost.

To tackle the first limitation, we propose a Gaussian process (GP) based kernel learning method to learn flexible kernel functions for a more accurate similarity measurement. We find that existing state-of-the-art methods \cite{min2021hypercorrelation,hong2021cost} mainly focus on the cost volume aggregation module and directly rely on the hypercorrelation tensors constructed by the cosine similarity. However, we argue that directly using the cosine similarity lacks flexibility and cannot faithfully reflect the semantic similarities between support and query pixels. As shown in Fig. \ref{compare} (a), the match between the cosine similarity map and the ground truth label is poor. %Thus, using the score maps computed by cosine similarity for hypercorrelation aggregation will result in limited performance and slow convergence speed.
To handle this problem, we design a hard example mining mechanism to dynamically sample hard samples to train the covariance kernel functions. As shown in Fig. \ref{compare} (b), using the learned covariance kernels can generate a more reasonable similarity map and greatly improve the convergence speed.

Another limitation is mainly caused by the cost volume aggregation module. Prior works have shown that a better cost volume aggregation method can attain superior few-shot segmentation performance. However, the existing 4D convolution-based method \cite{min2021hypercorrelation} lacks the ability to model longer distance relations between elements in the cost volume. VAT \cite{hong2021cost} proposes to combine 4D convolutions with 4D swin Transformers \cite{liu2021swin} for cost volume aggregation. Although outstanding performance can be achieved, it requires a notable GPU memory consumption and suffers from a slow convergence speed. Considering this, we propose a doubly deformable 4D Transformer based aggregation method, with deformable attention mechanisms on both the support and query dimensions of the 4D cost volume input. Compared with 4D convolutions, it can not only model longer-distance interactions between pixels owing to the Transformer network design, but also learn to selectively attend to more informative regions of the support and query information. %Compared with 4D swin transformer-based aggregation method \cite{hong2021cost}, the proposed plug-and-play DDT module requires much fewer GPU memory while maintain high segmentation performance.

To sum up, we propose a novel few-shot segmentation framework using doubly deformable aggregation of covariance matrices (DACM), which aggregates learnable covariance matrices with a doubly deformable 4D Transformer to predict segmentation results effectively. The proposed GP-based kernel learning can learn more accurate similarity measurements for cost volume generation. In what follows, the designed doubly deformable 4D Transformer can effectively and efficiently aggregate the multi-scale cost volume into the final segmentation result. Specifically, our contributions can be summarized as follows:

\begin{itemize}
    \item Towards a more accurate similarity measurement, a GP-based kernel learning method is proposed to learn flexible covariance kernel functions, with a novel hard example mining mechanism. To our knowledge, we are the first to use learnable covariance functions instead of cosine similarity for the few-shot segmentation task.
    
    \item Towards a more flexible cost volume aggregation, we propose a doubly deformable 4D Transformer (DDT) module, which utilizes deformable attention mechanisms on both the support and query dimensions of the 4D cost volume input. DDT can enhance representation learning by selectively attending to more informative regions.
    
    \item By combining these two modules, the proposed DACM method can attain new state-of-the-art performance on three public datasets with an extremely fast convergence speed. We also provide extensive visualization to better understand the proposed method.
\end{itemize}

\section{Related Work}
\subsection{Few-shot Segmentation}
Mainstream methods for few-shot segmentation can be roughly categorized into prototype-based methods \cite{yang2020prototype,wang2019panet,liu2020part,dong2018few} and correlation-based methods \cite{yang2020new,hu2019attention,zhang2019pyramid,wang2020few}. Prototype-based methods aim to generate a prototype representation \cite{snell2017prototypical} for each class based on the support sample, and then predict segmentation maps by measuring the distance between the prototype and the representations of the query image densely \cite{zhang2020sg,dong2018few,wang2019panet}. How to generate the prototype representation is the core of such methods. For example, a feature weighting and boosting model was proposed for prototype learning in \cite{nguyen2019feature}. Considering the limitations of using a single holistic representation for each class, a prototype learner was proposed in \cite{yang2020prototype} to learn multiple prototypes based on the support images and the corresponding object masks. Then, the query images and the prototypes were used to generate the final object masks. To capture diverse and fine-grained object characteristics, Part-aware Prototype Network (PPNet) was developed in \cite{liu2020part} to learn a set of part-aware prototypes for each semantic class. A prototype alignment regularization between support and query samples was proposed in Prototype Alignment Net-work (PANet) \cite{wang2019panet} to fully exploit semantic information from the support images and achieve better generalization on unseen classes. To handle the intra-class variations and inherent uncertainty, probabilistic latent variable-based prototype modeling was designed in \cite{sun2021attentional,wang2021variational}, which leveraged probabilistic modeling to enhance the segmentation performance.

As prototype-based methods suffer from the loss of spatial structure due to average pooling \cite{zhang2020sg}, correlation-based methods were proposed to model the pair-wise semantic correspondences densely between the support and query images. For instance, Attention-based Multi-Context Guiding (A-MCG) network \cite{hu2019attention} was proposed to fuse the multi-scale context features extracted from the support and query branch. Specifically, a residual attention module was introduced into A-MCG to enrich the context information. To retain the structure representations of segmentation data, attentive graph reasoning \cite{zhang2019pyramid} was exploited to transfer the class information from the support set to the query set, where element-to-element correspondences were captured at multiple semantic levels. Democratic Attention Networks (DAN) was introduced in \cite{wang2020few} for few-shot semantic segmentation, where the democratized graph attention mechanism was applied to establish a robust correspondence between support and query images. As proven in existing state-of-the-art methods, cost volume aggregation plays an important role in the few-shot segmentation task. To this end, 4D convolutions and Transformers were designed in \cite{min2021hypercorrelation,hong2021cost} and achieved highly competitive few-shot segmentation results.

\subsection{Gaussian Process}
Recently, the combination of the Gaussian process and deep neural networks has drawn more and more research attention due to the success of deep learning \cite{tossou2019adaptive,snell2020bayesian,patacchiola2020bayesian,wilson2016deep}. For instance, scalable deep kernels were introduced in \cite{wilson2016deep} to combine the non-parametric flexibility of GP and the structural properties of deep models. Furthermore, adaptive deep kernel learning \cite{tossou2019adaptive} was proposed to learn a family of kernels for numerous few-shot regression tasks, which enabled the determination of the appropriate kernel for specific tasks. Deep Kernel Transfer (DKT) \cite{patacchiola2020bayesian} was presented to learn a kernel that can be transferred to new tasks. Such a design can be implemented as a single optimizer and can provide reliable uncertainty measurement. GP has been introduced to few-shot segmentation in \cite{johnander2021dense}, where they proposed to learn the output space of the GP via a neural network that encoded the label mask. Although there is an attempt to incorporate GP, it is still in its infancy and far from fully exploiting the potential of GP for few-shot segmentation.

\section{Methodology}
\begin{figure}[t]
	\centering
	\includegraphics[width=0.95\textwidth]{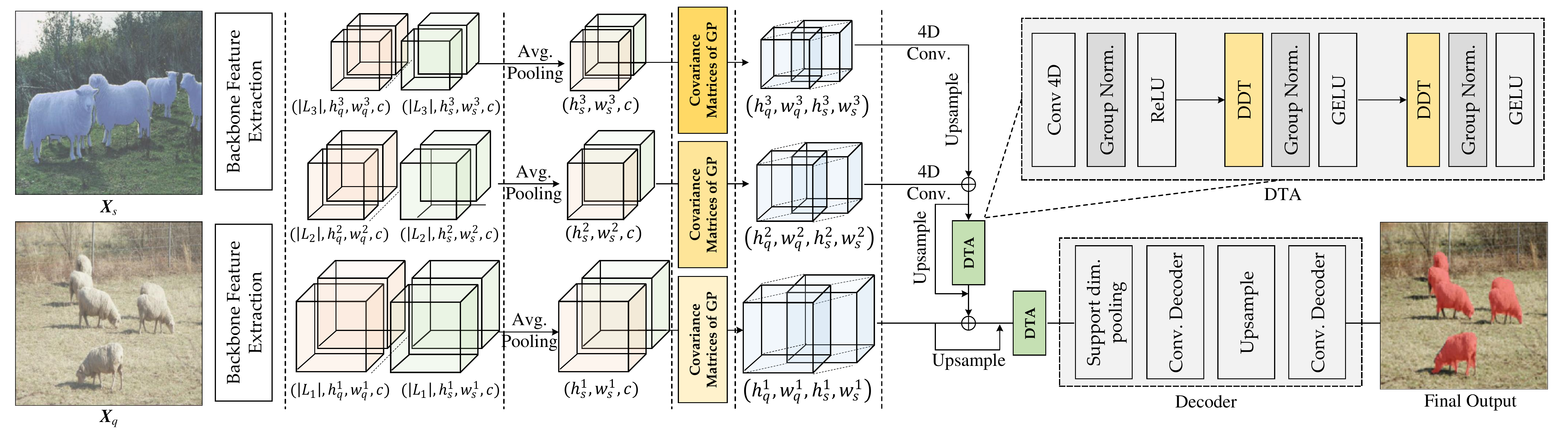}
	\caption{The whole architecture of our proposed DACM framework. Under a pyramidal design, DACM consists of three parts: 1) feature extraction by the backbone network; 2) cost volume generation based on covariance kernels; 3) cost volume aggregation using DTA modules and 4D convolutions \cite{min2021hypercorrelation}.}
	\label{arch}
\end{figure}

The whole architecture of the proposed DACM framework is illustrated in Fig. \ref{arch}. Given the input support and query images, multi-level deep features are first extracted by the fixed backbone network pre-trained on ImageNet \cite{russakovsky2015imagenet}. Three levels of deep features with different spatial resolutions form a pyramidal design. For each level, average pooling is used to aggregate multiple features into a single one. Then, three different GP models are trained for computing the covariance matrices (4D cost volume) for each level. Next, the proposed Deformable 4D Transformer Aggregator (DTA) is combined with the weight-sparsified 4D convolution for cost volume aggregation. Finally, a decoder is used to predict the final segmentation result for the input query image.

\subsection{Preliminaries: Gaussian Process}
\label{GP}

\begin{figure}[t]
	\centering
	\includegraphics[width=0.67\textwidth]{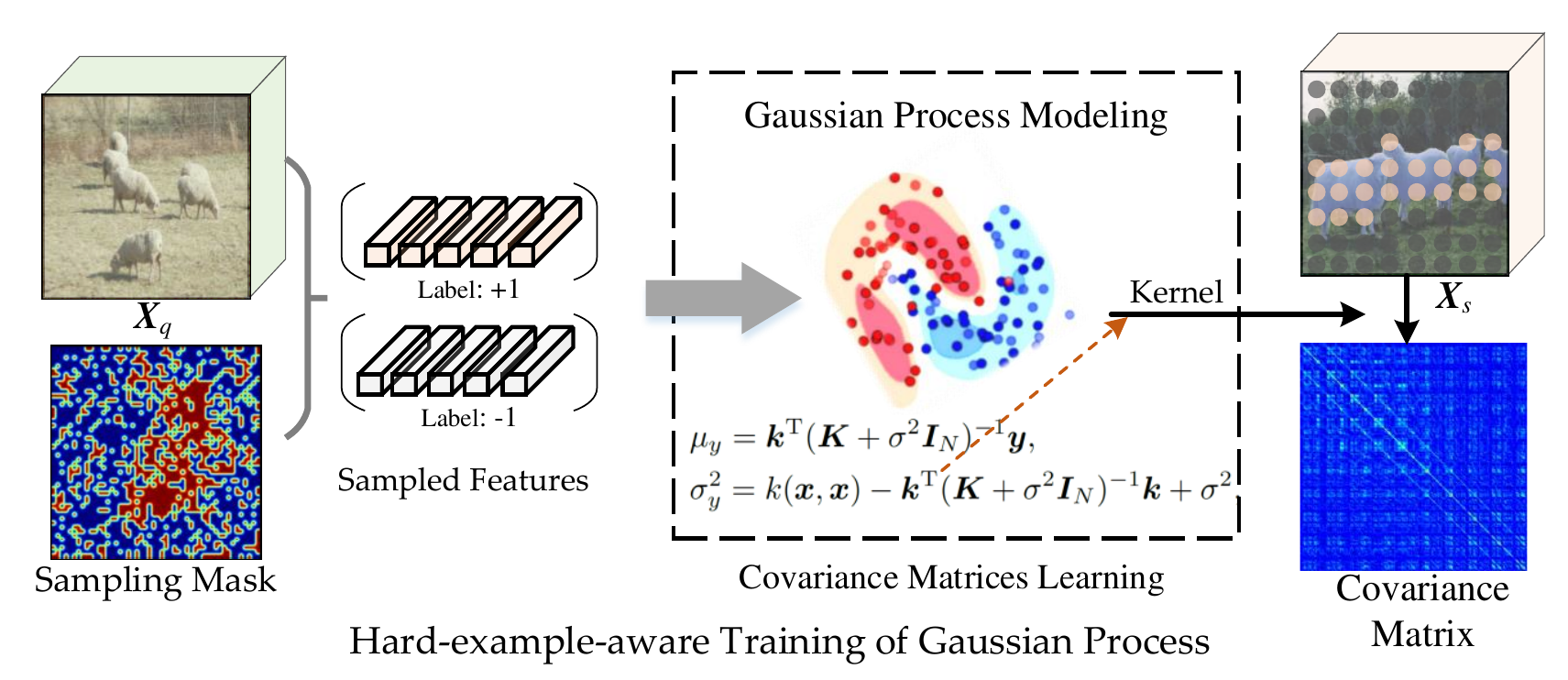}
	\caption{Illustration of the GP-based kernel learning process. Owing to Gaussian modeling, covariance functions can measure the similarity in a continuous feature space.}
	\label{subgp}
\end{figure}

% Why use GP for kernel learning. GP is a modeling the relationships
The covariance kernels of GP specify the statistical relationship between two points at the input space (\textit{e.g.}, support and query features). Namely, the learned covariance matrices can be naturally used to measure the correlations between support and query samples. Formally, the data set consists of $N$ samples of dimension $D$, \textit{i.e.}, $\{({x}_i,y_i)\rbrace_{i=1}^N$, where ${x}_i\in \mathbb{R}^D$ is a data point and $y_i$ is the corresponding label. The GP regression model assumes that the outputs $y_i$ can be regarded as certain deterministic latent function $f({x}_i)$ with zero-mean Gaussian noise $\varepsilon$, \textit{i.e.}, $y_i = f({x}_i)+\varepsilon$, where $\varepsilon\sim\mathcal{N}(0,\sigma^2)$. GP sets a zero-mean prior on $f$, with covariance $k({x}_i,{x}_j)$. The covariance function $k$ reflects the smoothness of $f$. The most widely-used covariance function is the Automatic Relevance Determination Squared Exponential (ARD SE), \textit{i.e.},
\begin{equation}
k({x}_i,{x}_j)=\sigma_0^2\exp\left\lbrace -\frac{1}{2}\sum_{d=1}^D\frac{(({x}_i)_d-({x}_j)_d)^2}{l_d^2}\right\rbrace,
\end{equation}
where $\sigma^2, \sigma_0^2,\left\lbrace l_d\right\rbrace_{d=1}^D$ are hyper-parameters. As shown in Fig. \ref{subgp}, after the optimization of hyper-parameters, we can compute the covariance matrices using the learned kernel $k$.

\subsection{Hard Example Mining-based GP Kernel Learning}
For few-shot segmentation, the whole dataset is split into the meta-training, meta-validation, and meta-testing subsets. Each \textit{episode} \cite{vinyals2016matching,shaban2017one} in the subsets consists of a support set $\mathcal{S}=\left\lbrace (\bm{I}_s^i,\bm{M}_s^i)\right\rbrace_{i=1}^K$ and a query image $\bm{I}_q$ of the same class, where $\bm{I}_s^i\in\mathbb{R}^{3\times H\times W}$ is the support image and $\bm{M}_s^i\in\{0,1\}^{H\times W}$ is its binary object mask for a certain class. $K$ is the number of the support images. We discuss the one-shot segmentation ($K=1$) case in the following sections.

Given the support image-mask pair $(\bm{I}_s,\bm{M}_s)$ and the query image-mask pair $(\bm{I}_q,\bm{M}_q)$ where $\bm{I}_s,\bm{I}_q\in \mathbb{R}^{3 \times H \times W}$ and $\bm{M}_s,\bm{M}_q\in\{0,1\}^{H\times W}$, a pretrained backbone is used to extract features from $\bm{I}_s$ and $\bm{I}_q$, and the feature maps at layer $l$ are denoted as $\bm{F}_s^l,\bm{F}_q^l\in \mathbb{R}^{c_l \times h_l \times w_l}$, where $c_l,h_l,w_l$ denote that the features are of channel $c_l$, height $h_l$ and width $w_l$. Note that we view the feature vectors at all spatial positions on feature maps of the query images as the training data. As shown in Fig. \ref{subgp}, our motivation is to dynamically select training samples from $\bm{F}_q^l$ for training the GP at layer $l$ with a hard example mining mechanism. To this end, we propose a hard example aware sampling strategy based on the similarity (cost volume) between support and query samples $\bm{F}_s^l$ and $\bm{F}_q^l$. The 4D covariance matrices $\mathcal{C}^l\in \mathbb{R}^{h_l\times w_l\times h_l \times w_l}$, \textit{i.e.}, the cost volume, can be computed with the GP using an initialized covariance kernel by
\begin{equation}
    \mathcal{C}^l(i,j)= \operatorname{ReLU} \left( k \left( \frac{\bm{F}_{q}^{l}(i)}{\left\|\bm{F}_{q}^{l}(i)\right\|}, \frac{\bm{F}_{s}^{l}(j)}{\left\|\bm{F}_{s}^{l}(j)\right\|} \right) \right),
\end{equation}
where $i$ and $j$ denote spatial positions of 2D feature maps. In what follows, we reshape the 4D cost volume into $\mathcal{C}^l \in \mathbb{R}^{h_l \times w_l \times h_l w_l}$ and sum up it along the third dimension. Then we can get a 2D similarity map $\mathcal{S}^l \in \mathbb{R}^{h_l \times w_l}$ of the query image. An example 2D similarity map can be found in Fig. \ref{compare}. Ideally, we expect the high values in $\mathcal{S}^l$ to totally lie in the ground truth mask. The negative positions with high similarity values in the 2D similarity map $\mathcal{S}^l$ can be viewed as hard examples. Thus, we utilize $\mathcal{S}^l$ to generate a probability map for sampling the training samples from $\bm{F}_q^l$. Specifically, we can obtain a 2D probability map for sampling training data by
\begin{align}
&\hat{\mathcal{\mathcal{S}}}^l=\mathcal{S}^l+\lambda \bm{M}_q^l, \label{e6}\\
&p^l=\frac{\hat{\mathcal{S}}^l-\operatorname{min}(\hat{\mathcal{S}}^l)}{\operatorname{max}(\hat{\mathcal{S}}^l)-\operatorname{min}(\hat{\mathcal{S}}^l)} \label{e7}\\
&t^l\sim \operatorname{Bernoulli }(p^l). \label{e8}
\end{align}
In Eq. \ref{e6}, we add $\mathcal{S}^l$ and the scaled query mask $\bm{M}_q^l$ with $\lambda$ to ensure that adequate number of positive samples can be selected for training at the early stage. Then, in Eq. \ref{e7}, we normalize $\mathcal{S}^l$ to represent the probability that the sample will be selected or not.
After the $\operatorname{softmax}$ operation, the 2D probability map $p^l\in [0,1]^{h_l\times w_l}$ can be obtained with each position representing the probability of being selected. Finally, a binary mask $t^l \in \{0,1\}^{h_l\times w_l}$ can be sampled from a $\operatorname{Bernoulli}$ distribution, where 0 means unselected and 1 means selected.

Such a sampling manner aims to sample query features which are likely to be incorrectly classified during the training stage. Then, we aim to use the sampled hard example-aware features on the query image to optimize the kernel. Suppose the sampled $N$ training data at feature layer $l$ is $\{(x^q_{j},y^q_{j})\}_{j=1}^N$ ($x^q_{j}\in\mathbb{R}^{c_l}$). Then these samples are used to learn the hyper-parameters $\left\lbrace \sigma^2,\sigma_0^2,\left\lbrace l_d\right\rbrace_{d=1}^D\right\rbrace$ of GP for computing the covariance matrices by maximizing the marginal likelihood. Please refer to \textbf{\S\,4} of the supplementary material for a more detailed definition.

\subsection{Doubly Deformable 4D Transformer for Cost Volume Aggregation}
%Considering that there are multiple cost volume tensors in a pyramidal design, directly using 4D convolutions or Transformers requires unacceptable amount of GPU memory and computational cost. Thus, \cite{min2021hypercorrelation} designed a weight-sparsified 4D convolution to remedy this problem.
%and VAT \cite{hong2021cost} extended window-based swin Transformer to the 4D case to decrease the computational cost. 
\begin{figure}[t]
	\centering
	\includegraphics[width=0.88\textwidth]{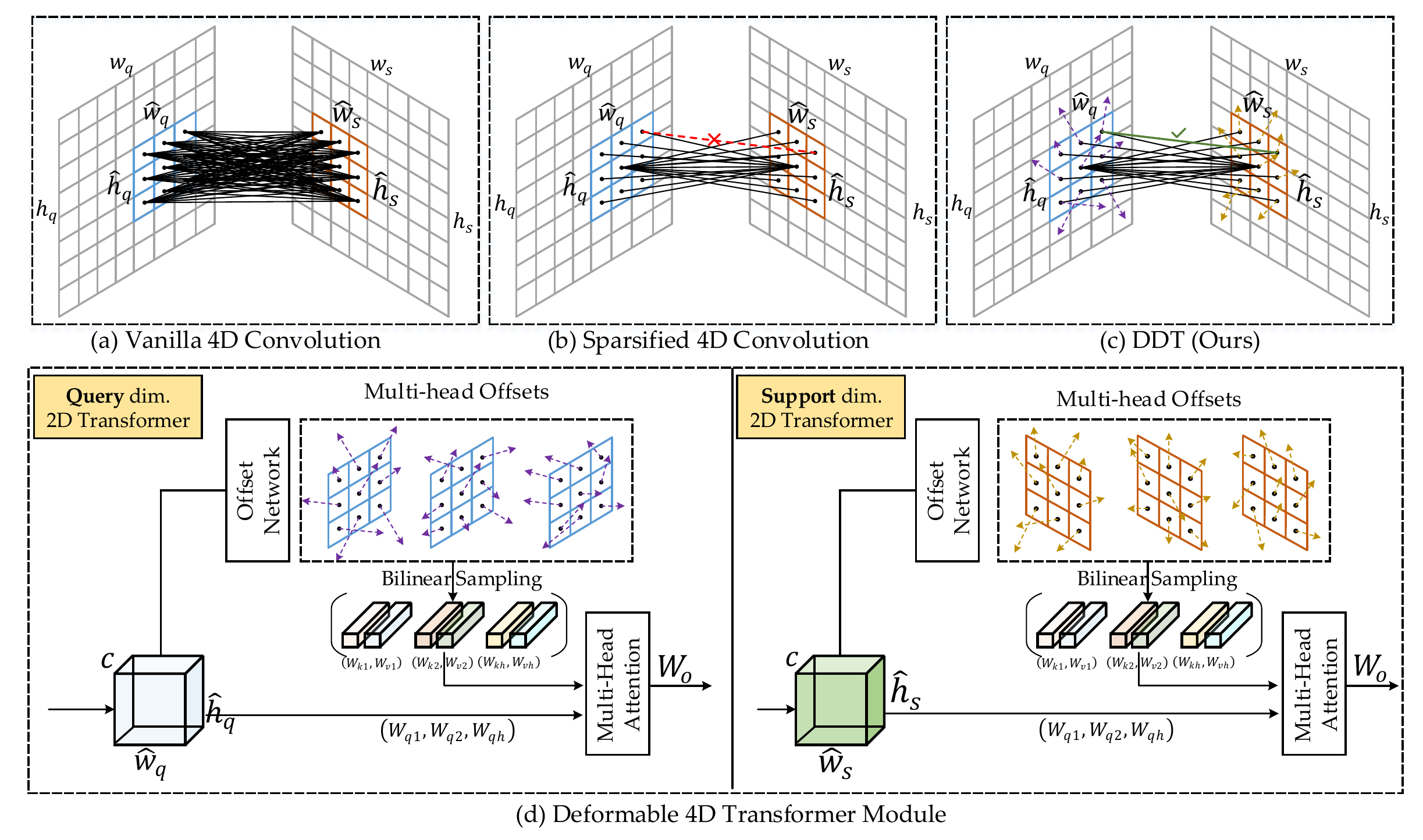}
	\caption{Illustration of the proposed DDT module. DDT utilizes doubly deformable attention mechanisms on both the support and query dimensions of the 4D cost volume. As shown in (b), the red dashed line indicates that the path is dropped by weight sparsification. While the dropped path can be recovered with the proposed DDT module if it is important, as presented in (c).}
	\label{d4dd}
\end{figure} 

To clearly describe the proposed DDT module, we will first introduce the sparsified 4D convolution proposed in \cite{min2021hypercorrelation}. As shown in Fig. \ref{d4dd}, weight-sparsified 4D convolution drops majority paths and only considers the activations at positions of either one of 2-dimensional centers. Formally, given 4D position $(\mathbf{u},\mathbf{u}{'})\in \mathbb{R}^4$, by only considering its neighbors adjacent to either $\mathbf{u}$ or $\mathbf{u}{'}$, the weight-sparsified 4D convolution can be approximated by two 2-dimensional convolutions performed on 2D slices of hypercorrelation tensor.
Different from prior works, the proposed DDT module aims to 1) utilize deformable attention mechanisms \cite{xia2022vision,zhu2020deformable} on both the support and query dimensions of the 4D input to learn more flexible aggregation; 2) provide the ability to model longer distance relations between elements of the cost volume. Specifically, we can formulate the DDT module for aggregating 4D tensor input $\mathcal{C}$ (we omit layer $l$ for simplicity) as follows:
\begin{equation}
\begin{aligned}
\mathrm{DDT}(\mathcal{C},\mathbf{u}, \mathbf{u}^{\prime})= \mathrm{SDT}\left(\mathcal{C}, (\mathbf{u}, \mathbf{p}^{\prime}) \right) + \mathrm{QDT}(\mathcal{C}, \left(\mathbf{p},  \mathbf{u}^{\prime}\right)),\\
\mathbf{p} := \mathcal{P}\left(\mathbf{u}\right), \,
\mathbf{p}^{\prime} := \mathcal{P}\left(\mathbf{u}^{\prime}\right),
\end{aligned}
\end{equation}
where $\mathbf{p}$ and $\mathbf{p}'$ denote the neighbour pixels centered at position $\mathbf{u}$ and $\mathbf{u}{'}$. $\mathrm{SDT}$ denotes the deformable Transformer on the support dimension (2D slice) of the cost volume. Similarly, $\mathrm{QDT}$ represents the deformable Transformer on the query dimension (2D slice) of $\mathcal{C}$. 

As $\mathrm{QDT}$ shares similar computation process with $\mathrm{SDT}$, we will only introduce the computation of $\mathrm{SDT}$ for the sake of simplicity. Suppose the 2D slice $\mathcal{C}(\mathbf{u},\mathbf{p}{'})\in \mathbb{R}^{c\times h \times w}$ of the cost volume is with channel dimension $c$, height $h$ and width $w$. Note that $c=1$ in this case. We first normalize the coordinates to the range $[-1,+1]$. $(-1,-1)$ indicates the top-left corner and $(+1,+1)$ represents the bottom-right corner. Then, an offset network is utilized to generate $n$ offset maps $\{\Delta \mathbf{p}_i{'}\}_{i=1}^n$ using two convolution layers. Each offset map has the spatial dimension $(h,w)$ and is responsible for a head in the multi-head attention module. By adding the original coordinates $\mathbf{p}{'}$ and $\Delta \mathbf{p}{'}$, we can obtain the shifted positions for the 2D slice input. In what follows, the differentiable bilinear sampling is used to obtain vectors $\tilde{x}\in \mathbb{R}^{c\times h \times w}$ at the shifted positions by:
\begin{equation}
\tilde{x}=\phi (\mathcal{C}(\mathbf{u},:), \mathbf{p}^{\prime}+\Delta \mathbf{p}^{\prime}).
\end{equation}
Then, we project the sample feature vectors into keys $k \in \mathbb{R}^{c{'}\times hw}$ and values $v \in \mathbb{R}^{c{'}\times hw}$ using learnable ${W_k}$ and ${W_v}$ as follows:
$k=\tilde{x} W_{k}$, $v=\tilde{x} W_{v}$. Where $c{'}$ is the projected channel dimension. Next, the query tokens $\mathcal{C}\left(\mathbf{u}, \mathbf{p}^{\prime} \right)$ in the query features are projected to queries $q\in \mathbb{R}^{c{'}\times hw}$. For each head of the multi-head attention, we can compute the output $z \in \mathbb{R}^{c{'}\times h \times w}$ using the self-attention mechanism as follows: 
\begin{equation}
q=\mathcal{C}\left(\mathbf{u}, \mathbf{p}^{\prime} \right) W_{q}, \, \, z=\operatorname{softmax} \left(q k^\mathrm{T} / \sqrt{d} \right) v,
\end{equation}
where $d$ denotes the dimension of each head, $W_{q}$ is a learnable matrix.
The whole computation process of DDT is illustrated in Fig. \ref{d4dd}. To sum up, the proposed light-weight, plug-and-play DDT module utilizes flexible deformable modeling to compensate the dropped informative activations in weight-sparsified 4D convolution. It can also enjoy the long-distance interactions modeling brought by Transformers.
\section{Experiments}

\subsection{Experiment Settings}

\subsubsection{Datasets}

Three few-shot segmentation datasets are exploited to evaluate the proposed method: PASCAL-$5^i$ \cite{shaban2017one}, COCO-$20^i$ \cite{lin2014microsoft}, and FSS-1000 \cite{li2020fss}. PASCAL-$5^i$ is re-created from the data in PASCAL VOC 2012 \cite{everingham2015pascal} and the augmented mask annotations from \cite{hariharan2014simultaneous}. PASCAL-$5^i$ contains 20 object classes and is split into 4 folds. COCO-$20^i$ consists of 80 object classes and is also split into 4 folds. Following prior experimental settings \cite{liu2020part,wang2020few,nguyen2019feature,tian2020prior} on PASCAL-$5^i$ and COCO-$20^i$ datasets, for each fold $i$, the data in the rest folds are used for training, and 1,000 episodes randomly sampled from the fold $i$ are used for evaluation. FSS-1000 contains 1,000 classes and is split into 520, 240, and 240 classes for training, validation, and testing, respectively.
\begin{figure}[tbb]
	\centering
	\includegraphics[width=0.92\textwidth]{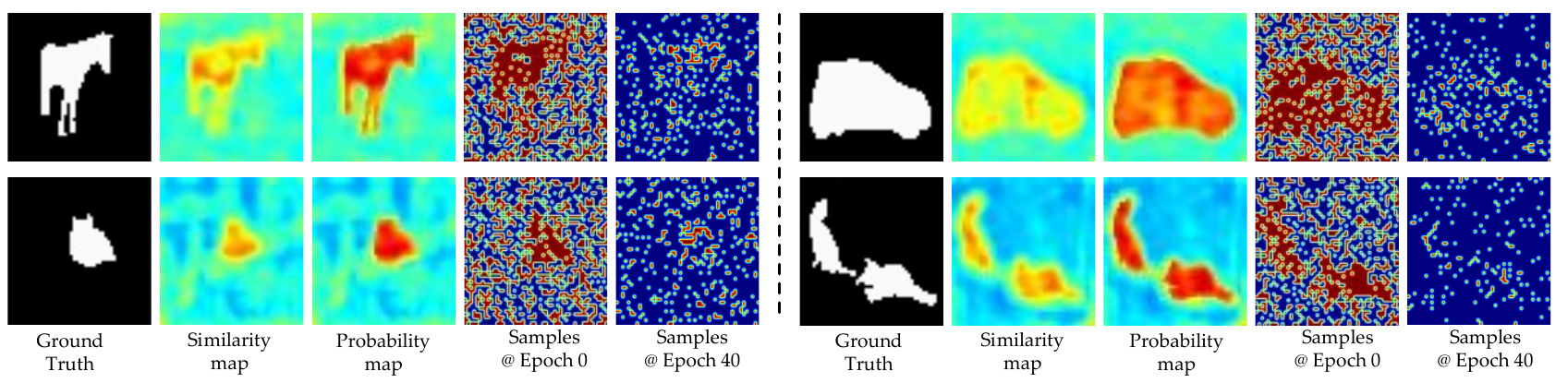}
	\caption{Some qualitative examples of the proposed hard example-aware sampling strategy for training the GP models.}
	\label{sampling_vis}
\end{figure} 
\subsubsection{Evaluation Metrics}
Mean intersection over union (mIoU) and foreground-background IoU (FB-IoU) are reported for comparisons. mIoU is the average of IoU over all classes in a fold, \textit{i.e.}, $\operatorname{mIoU}=\frac{1}{|\mathcal{C}_{test}|}\sum_{c\in\mathcal{C}}\operatorname{IoU}_c$. As for FB-IoU, the object classes are not considered in this metric. The average of foreground IoU and background IoU is computed, \textit{i.e.}, $\operatorname{FB-IoU}=\frac{1}{2}(\operatorname{IoU}_F+\operatorname{IoU}_B)$. Compared with FB-IoU, mIoU can better reflect the generalization ability of segmentation methods to unseen classes.

\subsubsection{Implementation Details}
The proposed method is implemented using PyTorch \cite{paszke2019pytorch} and GPyTorch \cite{gardner2018gpytorch}. VGG16 \cite{simonyan2014very}, ResNet50 and ResNet101 \cite{he2016deep} pre-trained on ImageNet are adopted as the backbone networks. To construct pyramidal features, three levels of features with resolutions of $50\times 50$, $25\times 25$, $13\times 13$ ($12\times 12$ for VGG backbone) are used. The GP models and cost volume aggregation models are trained in an end-to-end manner. Adam optimizer \cite{kingma2014adam} is used for the optimization of the covariance matrices aggregation part as well as the GP models. The initial learning rate for the covariance matrices aggregation part is set to $1e-3$, while the learning rate for three GP models is set to $1e-2$. Only \textbf{50 epochs} are used for training the proposed DACM on all three datasets. It is worth noting that 50 epochs are significantly less than existing methods that usually require more than 200 epochs for model training.

\begin{table}[tbp]
\caption{Performance comparisons on PASCAL-5$^i$ dataset in mIoU and FB-IoU.}
\label{pascal}
\center
\resizebox{0.928\textwidth}{!}{\begin{tabular}{clccccccccccccc}
\hline
\multirow{2}{*}{Backbone} & \multirow{2}{*}{Methods} & \multicolumn{6}{c}{1-shot Segmentation} & \multicolumn{6}{c}{5-shot Segmentation} & \multirow{2}{*}{\begin{tabular}[c]{@{}c@{}}\#Learnable\\ Params\end{tabular}} \\
 &  & 5$^0$ & 5$^1$ & 5$^2$ & 5$^3$ & Mean & FB-IoU & 5$^0$ & 5$^1$ & 5$^2$ & 5$^3$ & Mean & FB-IoU &  \\ \hline
\multirow{6}{*}{VGG16 \cite{simonyan2014very}} & co-FCN \cite{rakelly2018conditional} & 36.7 & 50.6 & 44.9 & 32.4 & 41.1 & 60.1 & 37.5 & 50.0 & 44.1 & 33.9 & 41.4 & 60.2 & 34.2M \\
 & AMP-2 \cite{siam2019adaptive}& 41.9 & 50.2 & 46.7 & 34.7 & 43.4 & 61.9 & 40.3 & 55.3 & 49.9 & 40.1 & 46.4 & 62.1 & 15.8M \\
 & PANet \cite{wang2019panet} & 42.3 & 58.0 & 51.1 & 41.2 & 48.1 & 66.5 & 51.8 & 64.6 & 59.8 & 46.5 & 55.7 & 70.7 & 14.7M \\
 & PFENet \cite{tian2020prior} & 56.9 & \textbf{68.2} & 54.4 & 52.4 & 58.0 & 72.0 & 59.0 & \underline{69.1} & 54.8 & 52.9 & 59.0 & 72.3 & 10.4M \\ 
 & HSNet \cite{min2021hypercorrelation}& \underline{59.6} & 65.7 & \underline{59.6} & \underline{54.0} & \underline{59.7} & \underline{73.4} & \underline{64.9} & 69.0 & \underline{64.1} & \underline{58.6} & \underline{64.1} & \underline{76.6} & \textbf{2.6M}\\\cline{2-15}
& DACM (Ours) &\textbf{61.8} & \underline{67.8} & \textbf{61.4} & \textbf{56.3}  & \textbf{61.8} & \textbf{75.5} & \textbf{66.1} & \textbf{70.6} & \textbf{65.8} & \textbf{60.2} & \textbf{65.7} & \textbf{77.8} & \underline{3.0M} \\\cline{2-15} \hline

\multirow{8}{*}{ResNet50 \cite{he2016deep}}& PPNet \cite{liu2020part}& 48.6 & 60.6 & 55.7 & 46.5 & 52.8 & 69.2 & 58.9 & 68.3 & 66.8 & 58.0 & 63.0 & 75.8 & 31.5M \\
 & PFENet \cite{tian2020prior}& 61.7 & 69.5 & 55.4 & 56.3 & 60.8 & 73.3 & 63.1 & 70.7 & 55.8 & 57.9 & 61.9 & 73.9 & 10.8M \\
 & RePRI \cite{boudiaf2021few}& 59.8 & 68.3 & 62.1 & 48.5 & 59.7 & --- & 64.6 & 71.4 & \textbf{71.1} & 59.3 & 66.6 & --- & --- \\ 
 &HSNet \cite{min2021hypercorrelation}& 64.3 & 70.7 & 60.3 & 60.5 & 64.0 & 76.7 & 70.3 & 73.2 & 67.4 & 67.1 & 69.5 & 80.6 & \textbf{2.6M}\\
 &CyCTR \cite{zhang2021few}& \underline{67.8} & \underline{72.8} & 58.0 & 58.0 & 64.2 & --- & 71.1 & 73.2 & 60.5 & 57.5 & 65.6 & --- & ---\\ 
 & VAT \cite{hong2021cost} & 67.6 &71.2 &\underline{62.3} &60.1 &65.3 &77.4 &\underline{72.4} &73.6 &68.6 &65.7 &70.0 &80.9 & 3.2M\\ \cline{2-15} 
 & DACM (Ours) &66.5 & 72.6 & 62.2 & \underline{61.3} & \underline{65.7}& \underline{77.8} & \underline{72.4} & \underline{73.7} & 69.1 & \textbf{68.4} & \underline{70.9} & \underline{81.3} & \underline{3.0M} \\
 & DACM (VAT) &\textbf{68.4} & \textbf{73.1} & \textbf{63.5} & \textbf{62.2} & \textbf{66.8} & \textbf{78.6} & \textbf{73.8} & \textbf{74.7} & \underline{70.3} & \underline{68.1} & \textbf{71.7} & \textbf{81.7} & 3.3M \\ \hline

\multirow{10}{*}{ResNet101 \cite{he2016deep}} & FWB \cite{nguyen2019feature}& 51.3 & 64.5 & 56.7 & 52.2 & 56.2 & --- & 54.8 & 67.4 & 62.2 & 55.3 & 59.9 & --- & 43.0M \\
 & PPNet \cite{liu2020part}& 52.7 & 62.8 & 57.4 & 47.7 & 55.2 & 70.9 & 60.3 & 70.0 & 69.4 & 60.7 & 65.1 & 77.5 & 50.5M \\
 & DAN \cite{wang2020few}& 54.7 & 68.6 & 57.8 & 51.6 & 58.2 & 71.9 & 57.9 & 69.0 & 60.1 & 54.9 & 60.5 & 72.3 & --- \\
 & PFENet \cite{tian2020prior}& 60.5 & 69.4 & 54.4 & 55.9 & 60.1 & 72.9 & 62.8 & 70.4 & 54.9 & 57.6 & 61.4 & 73.5 & 10.8M \\
 & RePRI \cite{boudiaf2021few}& 59.6 & 68.6 & 62.2 & 47.2 & 59.4 & --- & 66.2 & 71.4 & 67.0 & 57.7 & 65.6 & --- & --- \\ 
 &HSNet \cite{min2021hypercorrelation} & 67.3 & 72.3 & 62.0 & 63.1 & 66.2 & 77.6 & 71.8 & 74.4 & 67.0 & 68.3 & 70.4 & 80.6 & \textbf{2.6M}\\
  &CyCTR \cite{zhang2021few} &\underline{69.3} &72.7 &56.5 &58.6 &64.3 &72.9 &\underline{73.5} &74.0 &58.6 &60.2 &66.6 &75.0 &--- \\
  &VAT \cite{hong2021cost} &68.4 &72.5 &\underline{64.8} &\underline{64.2} &\underline{67.5} &78.8 &73.3 &75.2 &\underline{68.4} &\underline{69.5} &\underline{71.6} &\underline{82.0} & 3.3M \\\cline{2-15}
& DACM (Ours) &68.7 & \underline{73.5} & 63.4 & \underline{64.2}  & \underline{67.5} & \underline{78.9} & 72.7 & \underline{75.3} & 68.3 & 69.2 & 71.4 & 81.5 & \underline{3.1M} \\
& DACM (VAT)  &\textbf{69.9} & \textbf{74.1} & \textbf{66.2} & \textbf{66.0}  & \textbf{69.1} & \textbf{79.4} & \textbf{74.2} & \textbf{76.4} & \textbf{71.1} & \textbf{71.6} & \textbf{73.3} & \textbf{83.1} & 3.4M \\\cline{2-15} \hline
\end{tabular}}
\end{table}
\subsection{Results and Analysis}
\subsubsection{PASCAL-$5^i$}
\begin{figure}[t]
	\centering
	\includegraphics[width=0.82\textwidth]{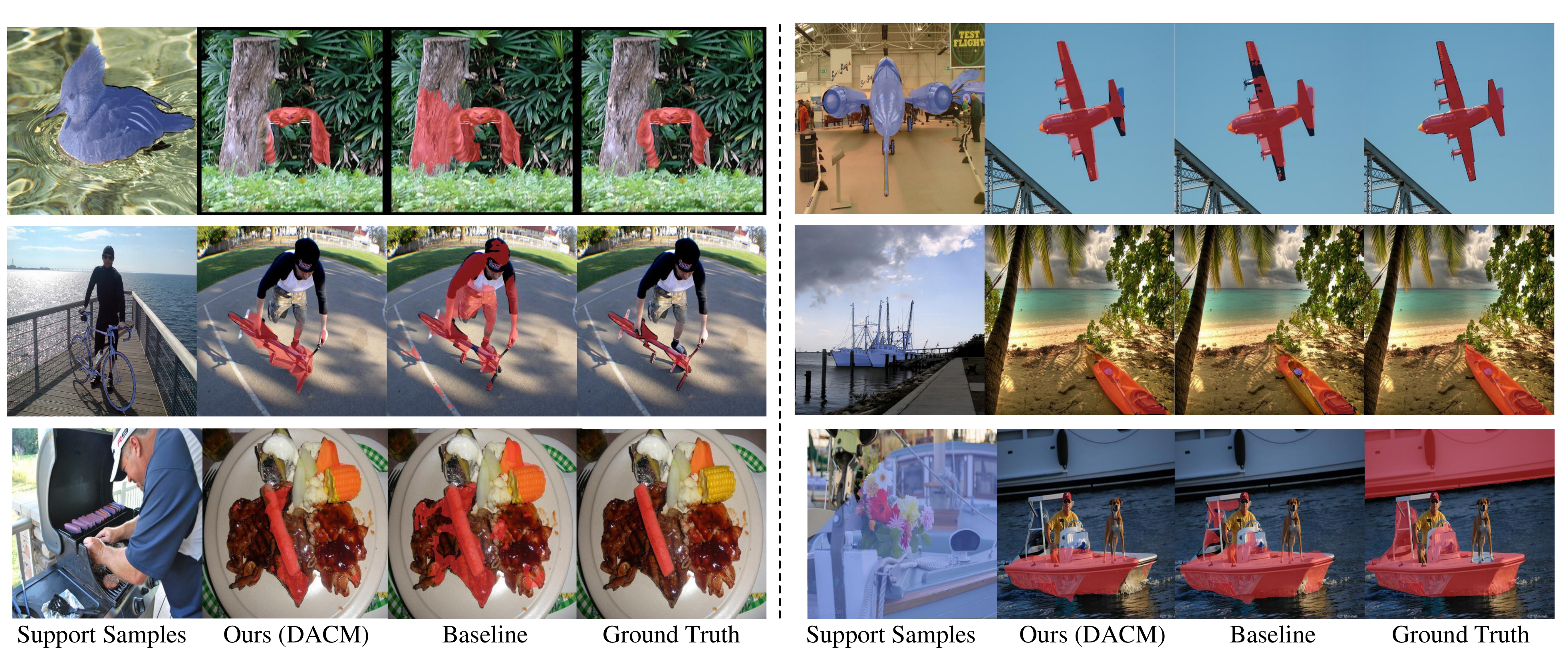}
	\caption{Some visualization examples of the proposed method for few-shot segmentation. We compare DACM with the baseline (HSNet) method.
}
	\label{R1}
\end{figure}

We compare our method with existing state-of-the-art methods. Table \ref{pascal} presents the 1-shot and 5-shot segmentation results. ``DACM (Ours)'' demotes the model presented in Fig. \ref{arch}. Although DACM is trained with only \textbf{50 epochs}, it can still significantly outperform existing state-of-the-art models \cite{min2021hypercorrelation,hong2021cost} with three different backbones. The comparison results demonstrate that the proposed GP-based kernel learning and DDT module are beneficial for few-shot segmentation task. In addition, we also visualize the sampled locations (described in Eq. \ref{e8}) during the GP kernel training in Fig. \ref{sampling_vis} for a better understanding of the designed hard example-aware sampling strategy. It can be seen that hard examples are selected at the later stage of the training process. Some comparative visualization results of ``DACM (Ours)'' model on this dataset are presented in Fig. \ref{R1}.

The proposed GP-based kernel learning and DDT module can be plugged in any cost volume aggregation-based model. ``DACM+VAT" denotes the combination of DACM with VAT by 1) replacing the cosine similarity-based cost volume with our covariance matrices; 2) replacing the last 4D convolution layer with our DDT.  We can see that the combined method ``DACM (VAT)'' can achieve better segmentation performance than others. This demonstrates the superiority of the proposed framework. 

% \usepackage{multirow}
\begin{table}[tbp]
\caption{Performance comparisons on COCO-20$^i$ dataset in mIoU and FB-IoU.}
\label{coco}
\center
\resizebox{0.88\textwidth}{!}{\begin{tabular}{clcccccccccccc}
\hline
\multirow{2}{*}{Backbone} & \multirow{2}{*}{Methods} & \multicolumn{6}{c}{1-shot Segmentation} & \multicolumn{6}{c}{5-shot Segmentation} \\
 &  & 5$^0$ & 5$^1$ & 5$^2$ & 5$^3$ & Mean & FB-IoU & 5$^0$ & 5$^1$ & 5$^2$ & 5$^3$ & Mean & FB-IoU \\ \hline
\multirow{9}{*}{ResNet50 \cite{he2016deep}}
 & PMM \cite{yang2020prototype}& 29.3 & 34.8 & 27.1 & 27.3 & 29.6 & --- & 33.0 & 40.6 & 30.3 & 33.3 & 34.3 & --- \\
 & RPMM \cite{yang2020prototype}& 29.5 & 36.8 & 28.9 & 27.0 & 30.6 & --- & 33.8 & 42.0 & 33.0 & 33.3 & 35.5 & --- \\
 & PFENet \cite{tian2020prior}& 36.5 & 38.6 & 35.0 & 33.8 & 35.8 & --- & 36.5 & 43.3 & 38.0 & 38.4 & 39.0 & --- \\
 & RePRI \cite{boudiaf2021few}& 32.0 & 38.7 & 32.7 & 33.1 & 34.1 & --- & 39.3 & 45.4 & 39.7 & 41.8 & 41.6 & --- \\
 & HSNet \cite{min2021hypercorrelation}& 36.3 & 43.1 & 38.7 & 38.7 & 39.2 & 68.2 & 43.3 & 51.3 & 48.2 & 45.0 & 46.9 & 70.7 \\ 
 & CyCTR \cite{zhang2021few}&38.9 &43.0 &39.6 &39.8 &40.3 &---   &41.1 &48.9 &45.2 &\underline{47.0} &45.6 &--- \\
 & VAT \cite{hong2021cost} &\underline{39.0} &43.8 &\underline{42.6} &39.7 &\underline{41.3} &68.8 &44.1 &51.1 &\underline{50.2} &46.1 &47.9  &\underline{72.4} \\ \cline{2-14} 
& DACM (Ours) &37.5 & \underline{44.3} & 40.6 & \underline{40.1} & 40.6 & \underline{68.9} & \underline{44.6} & \underline{52.0} & 49.2 & 46.4 & \underline{48.1}& 71.6 \\ 
& DACM (VAT) &\textbf{41.2} & \textbf{45.2} & \textbf{44.1} & \textbf{41.3}  & \textbf{43.0} & \textbf{69.4} & \textbf{45.2} & \textbf{52.2} & \textbf{51.5} & \textbf{47.7} & \textbf{49.2} &\textbf{72.9} \\ \hline
\end{tabular}}
\end{table}
\subsubsection{COCO-$20^i$}
The comparison results on the COCO-$20^i$ dataset are reported in Table \ref{coco}. The results reveal that ``DACM (Ours)'' clearly outperform the HSNet baseline under both 1-shot and 5-shot settings. ``DACM (VAT)'' model achieves the best results compared with existing methods. COCO-$20^i$ is a more difficult few-shot segmentation dataset. Basically, taking HSNet \cite{min2021hypercorrelation} as the baseline, DACM can achieve an clear improvement. It is worth noting that DACM requires less epochs for the model training, while it can still outperform VAT \cite{hong2021cost} and set new state-of-the-art results on the COCO-$20^i$ dataset.

\begin{table}[tbp]
\parbox{.45\linewidth}{
\caption{Performance Comparisons on the FSS-1000 dataset.}
\label{FssR}
\centering
\scalebox{0.88}{
\begin{tabular}{clcc}
\hline
\multirow{2}{*}{Backbone} & \multirow{2}{*}{Methods} & \multicolumn{2}{c}{mIoU} \\
 &  & 1-shot & 5-shot \\ \hline
\multirow{4}{*}{ResNet50 \cite{he2016deep}} 
& FSOT \cite{liu2021few}& 82.5 & 83.8 \\ 
& HSNet \cite{min2021hypercorrelation}& {85.5} & {87.8} \\ 
& VAT \cite{hong2021cost}& \underline{89.5} & \underline{90.3} \\ \cline{2-4} 
& DACM (Ours) & \textbf{90.7} & \textbf{91.6} \\ \hline
\multirow{4}{*}{ResNet101 \cite{he2016deep}} 
& DAN \cite{wang2020few}& 85.2 & 88.1 \\
& HSNet \cite{min2021hypercorrelation}& 86.5 & 88.5 \\ 
& VAT \cite{hong2021cost}& \underline{90.0} & \underline{90.6} \\ \cline{2-4} 
& DACM (Ours)  & \textbf{90.8}  & \textbf{91.7} \\ \hline
\end{tabular}}
}
\hfill
\parbox{.45\linewidth}{
\caption{Ablation study on PASCAL-$5^i$ dataset.}\label{abpas}
\centering
\scalebox{0.88}{
\begin{tabular}{clcc}
\hline
\multirow{2}{*}{Backbone} & \multirow{2}{*}{Methods} & \multicolumn{2}{c}{mIoU} \\
 &  & 1-shot & 5-shot \\ \hline
\multirow{6}{*}{VGG16 \cite{simonyan2014very}} 
 & Baseline \cite{min2021hypercorrelation} & 59.6 & 64.9 \\  \cline{2-4} 
 & + GP-KL & 60.8 & 65.2  \\
 & + DDT-1  & 61.2 & 65.7 \\
 & + DDT-2  & \textbf{61.8} & \textbf{66.1} \\
 & + DDT-3  & \underline{61.6} & \underline{65.8} \\ \hline
\end{tabular}}
}
\end{table}
\subsubsection{FSS-1000}
FSS-1000 is simpler than the other two datasets. The comparison results are shown in Table \ref{FssR}. Under both the 1-shot setting and 5-shot setting, DACM can obtain a clear new state-of-the-art performance in terms of mIoU. This indicates the effectiveness of our method. Comparison results on these three public datasets demonstrate the effectiveness of the proposed DACM method.

\begin{figure*}
	\centering
	\subfigure[]{
		\begin{minipage}[b]{0.22\textwidth}
			\includegraphics[width=1\textwidth]{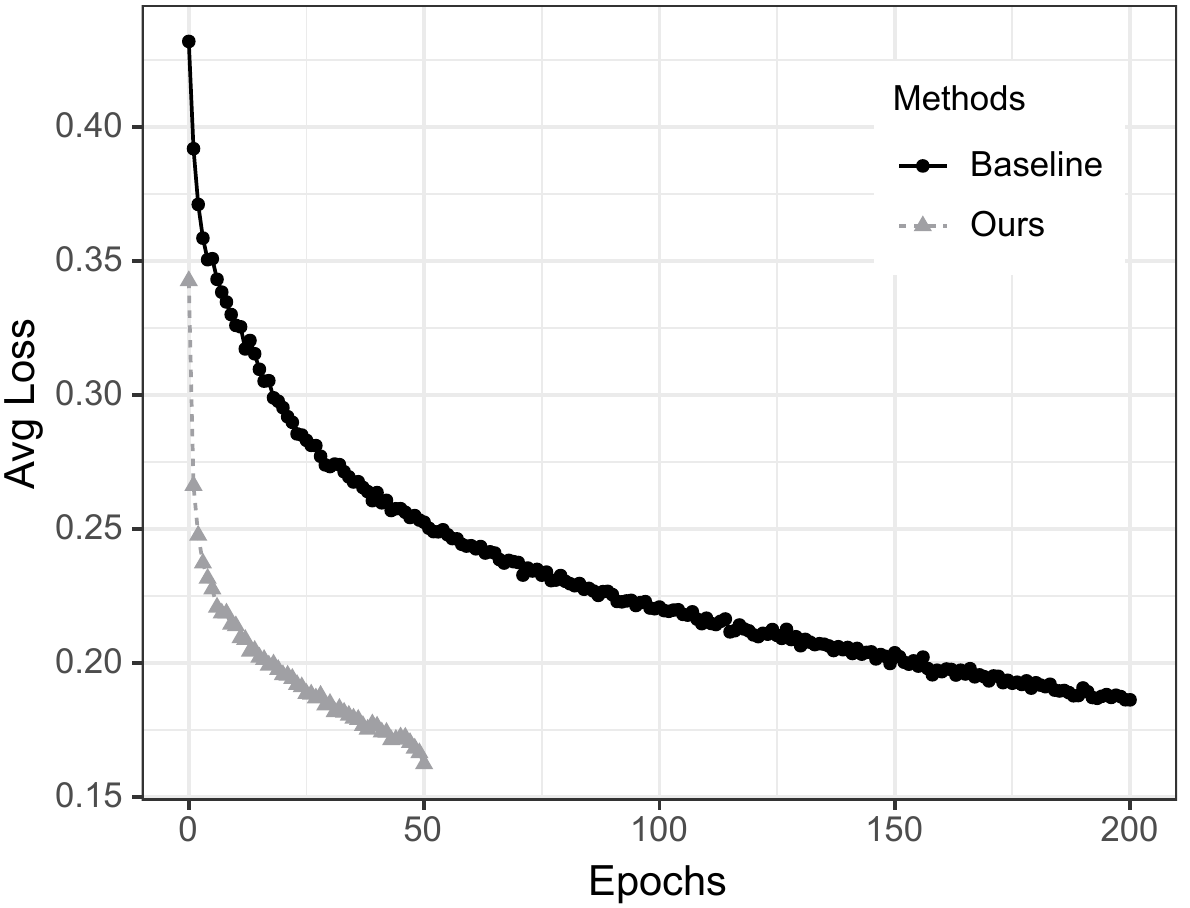} 
		\end{minipage}
		\label{fig:grid_4figs_1cap_4subcap_1}
	}
    	\subfigure[]{
    		\begin{minipage}[b]{0.22\textwidth}
   		 	\includegraphics[width=1\textwidth]{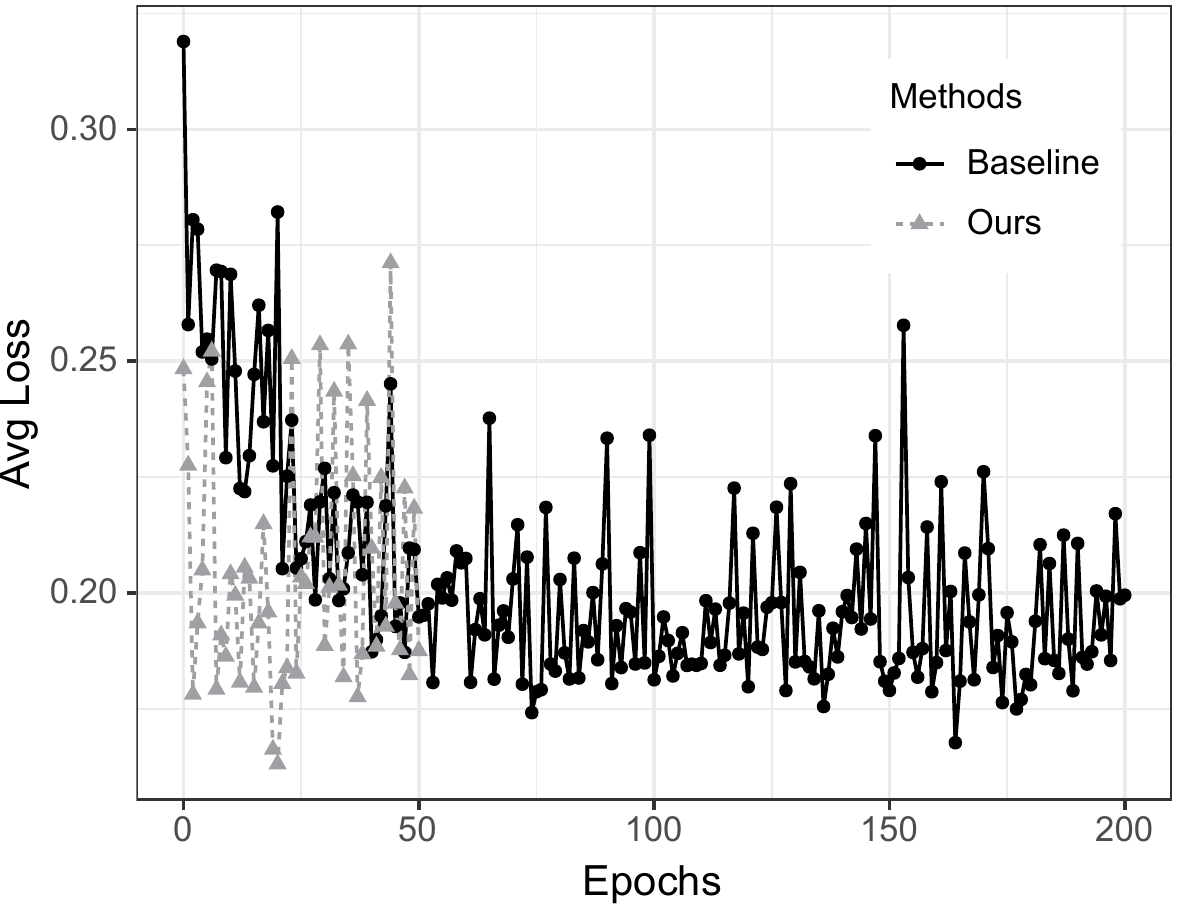}
    		\end{minipage}
		\label{fig:grid_4figs_1cap_4subcap_2}
    	}
	\subfigure[]{
		\begin{minipage}[b]{0.22\textwidth}
			\includegraphics[width=1\textwidth]{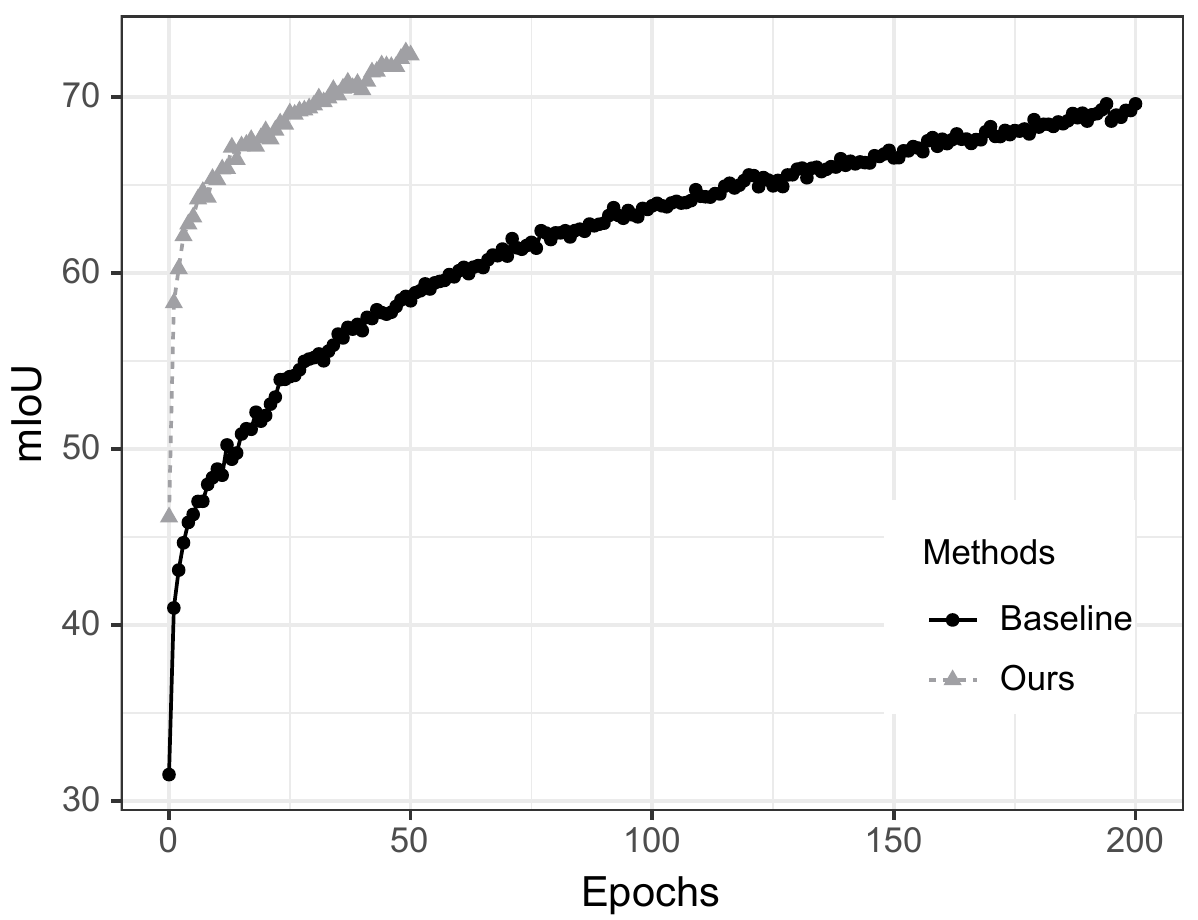} 
		\end{minipage}
		\label{fig:grid_4figs_1cap_4subcap_3}
	}
    	\subfigure[]{
    		\begin{minipage}[b]{0.22\textwidth}
		 	\includegraphics[width=1\textwidth]{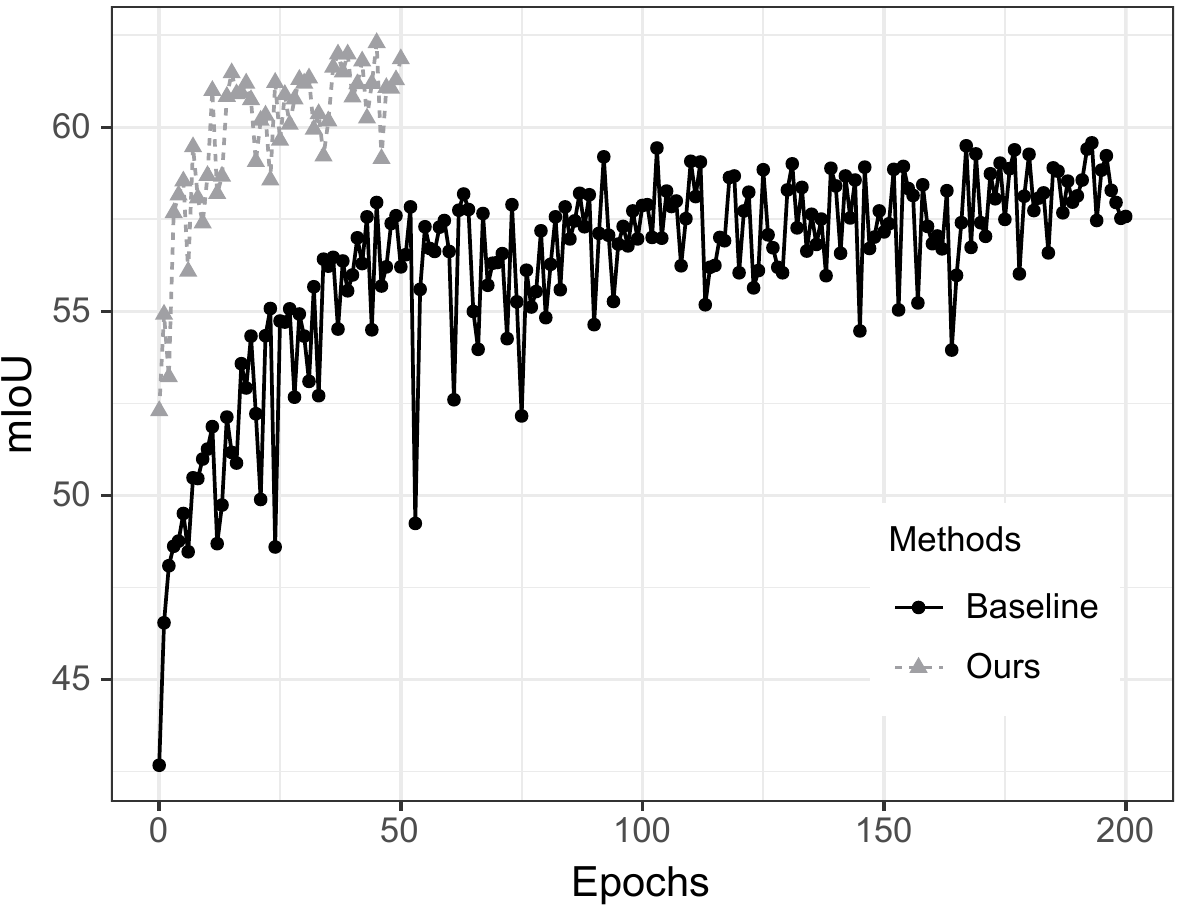}
    		\end{minipage}
		\label{fig:grid_4figs_1cap_4subcap_4}
    	}
	\centering
	\subfigure[]{
		\begin{minipage}[b]{0.22\textwidth}
			\includegraphics[width=1\textwidth]{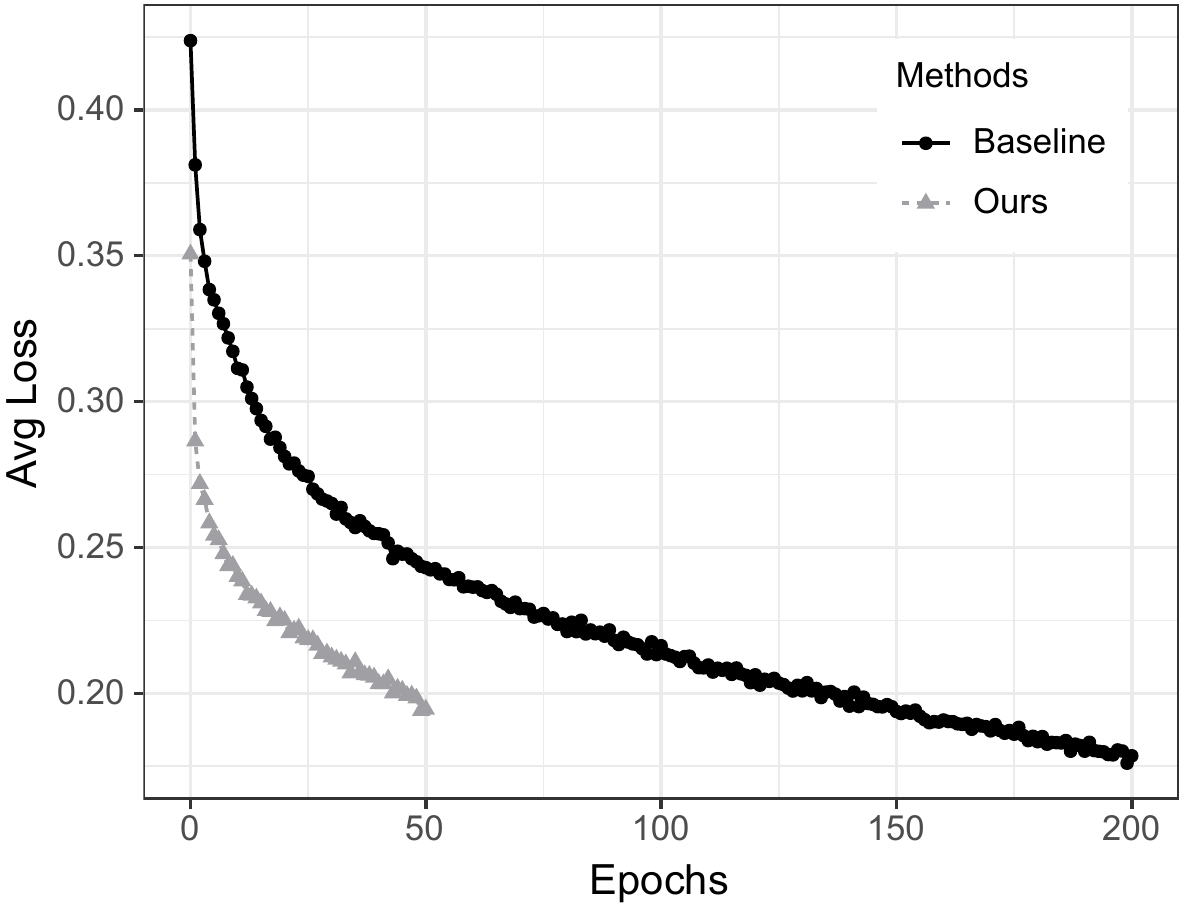} 
		\end{minipage}
		\label{fig:grid_4figs_1cap_4subcap_5}
	}
    	\subfigure[]{
    		\begin{minipage}[b]{0.22\textwidth}
   		 	\includegraphics[width=1\textwidth]{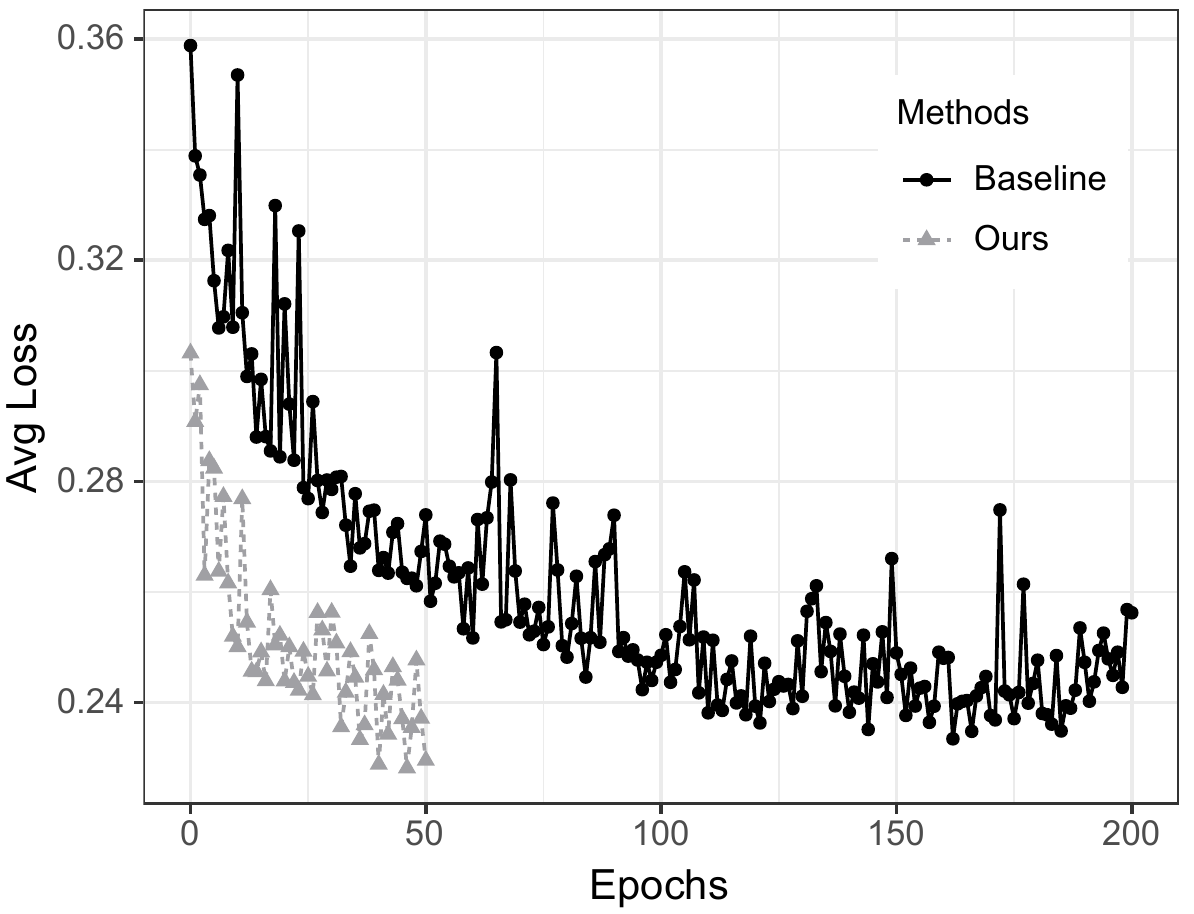}
    		\end{minipage}
		\label{fig:grid_4figs_1cap_4subcap_6}
    	}
	\subfigure[]{
		\begin{minipage}[b]{0.22\textwidth}
			\includegraphics[width=1\textwidth]{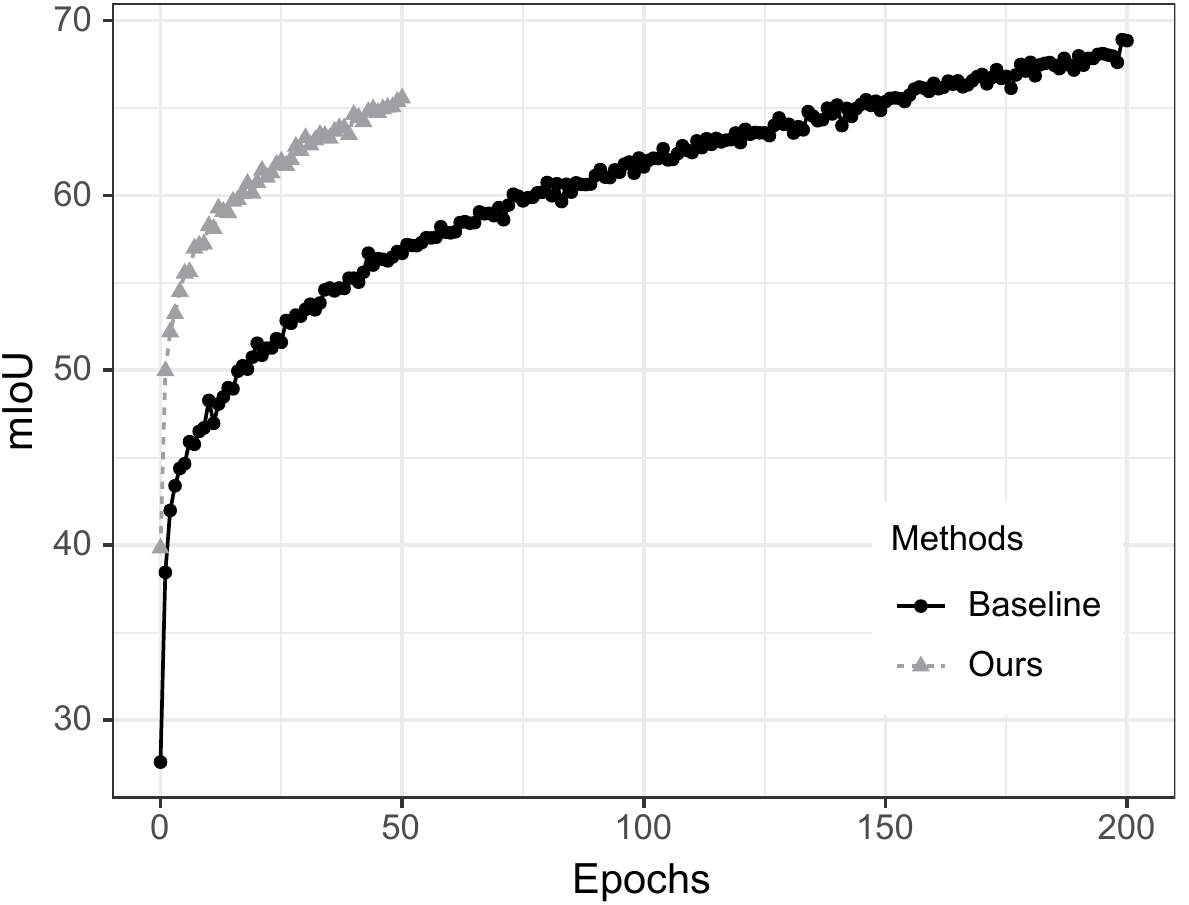} 
		\end{minipage}
		\label{fig:grid_4figs_1cap_4subcap_7}
	}
    	\subfigure[]{
    		\begin{minipage}[b]{0.22\textwidth}
		 	\includegraphics[width=1\textwidth]{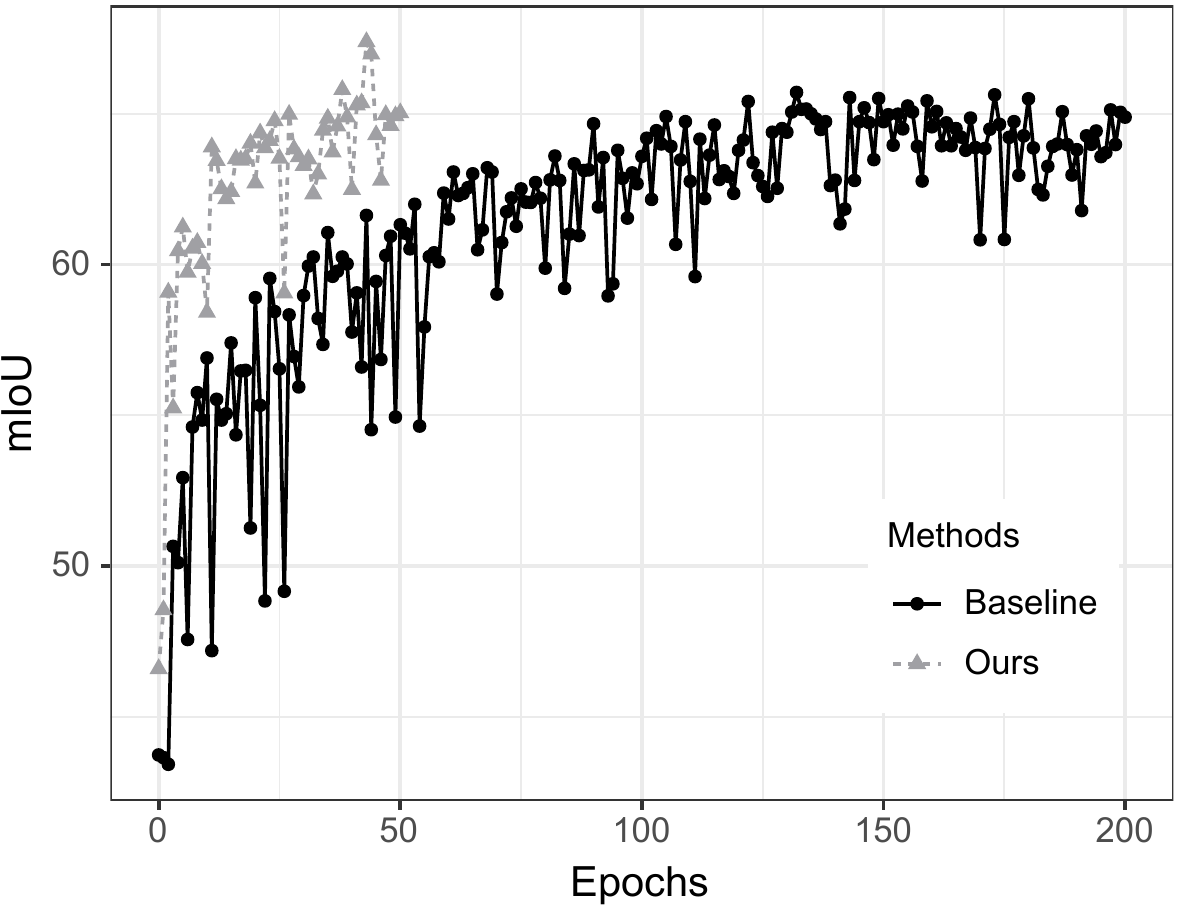}
    		\end{minipage}
		\label{fig:grid_4figs_1cap_4subcap_8}
    	}
	\caption{(a), (b), (e), and (f) show the training/validation loss curves on the fold 0 and fold 1 of PASCAL-$5i$ dataset. (c), (d), (g), and (h) present the training/validation mIoU curves.}
	\label{curves}
\end{figure*}
\subsection{Ablation Study and Analysis}
\subsubsection{Ablation Study}
To validate the effects of the proposed two modules, \textit{i.e.}, GP-based kernel learning (GP-KL) module and DDT module, we conduct ablation studies. In addition, we compare the performance of using 1, 2 and 3 DDT layers (DDT-1, DDT-2 and DDT-3) to further study the effects of the number of DDT layers. The mIoU results on the PASCAL-$5^i$ dataset are presented in Table \ref{abpas}. We can see that using 2 DDT layers can obtain the best result. The reason is that more DDT layers may bring the overfitting problem on several folds of the dataset. We also find that combining the GP-based kernel learning and DDT modules together results in the best few-shot segmentation performance.
\begin{figure}[t]
	\centering
	\includegraphics[width=0.92\textwidth]{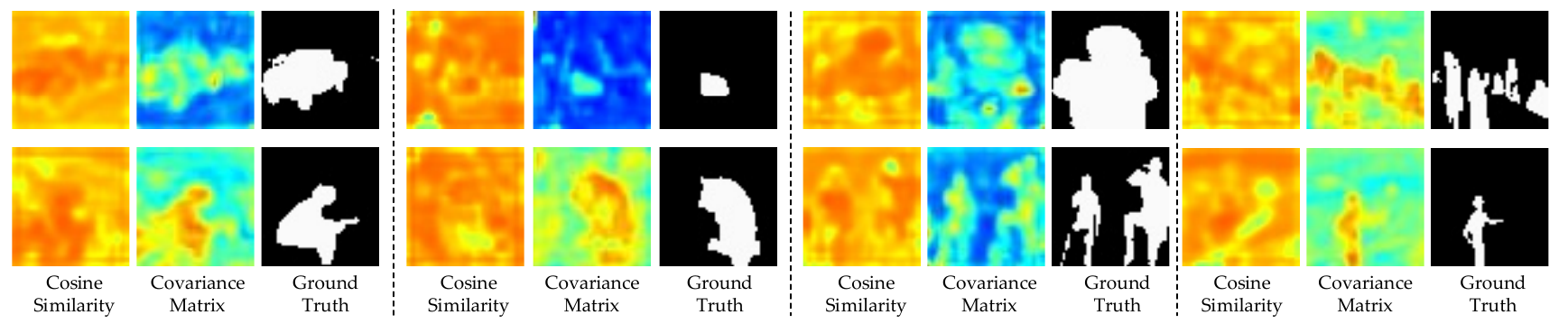}
	\caption{Visualization of some examples of the covariance matrices. Compared with the commonly used cosine similarity, our method can learn more reasonable similarity measurements for few-shot segmentation.}
	\label{simmap}
\end{figure} 
\begin{figure}[t]
	\centering
	\includegraphics[width=0.92\textwidth]{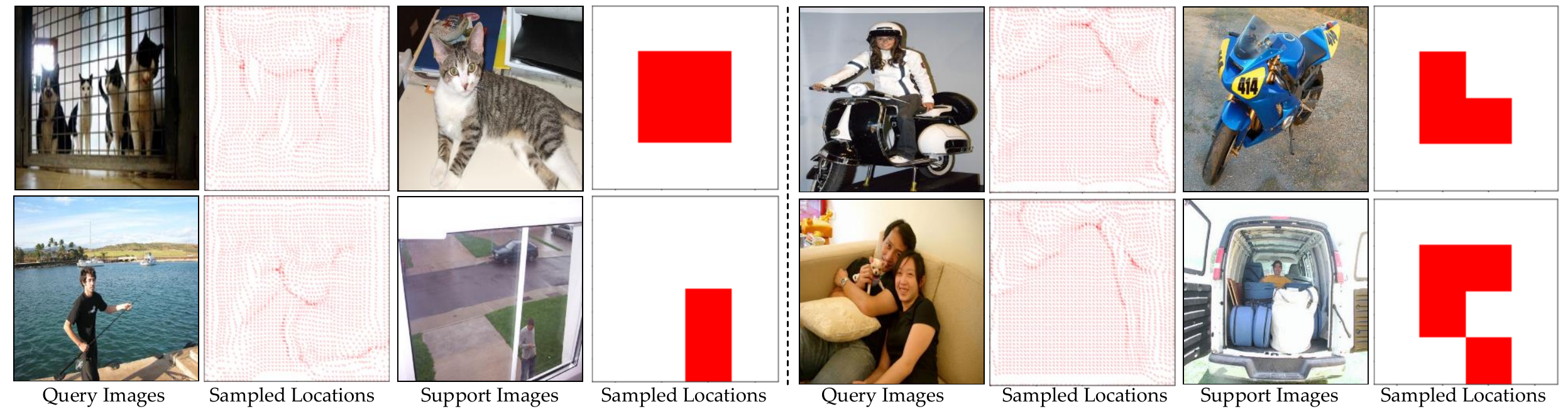}
	\caption{Visualization of some examples of the learned deformable offsets for both the query and support samples.}
	\label{vis_defor}
\end{figure} 
\subsubsection{Convergence Speed}
In Fig. \ref{curves}, we plot the loss curves and mIoU values during the training process on two folds of the PASCAL-$5^i$ dataset. Obviously, our DACM can converge much faster than the baseline method (HSNet). Although trained with only 50 epochs, DACM can obtain much better training and validation mIoU performance than the baseline method. This clearly demonstrates the effectiveness of the proposed modules.
\subsubsection{Covariance Matrices Visualization}
To better understand the proposed DACM method, we further visualize and compare the learned covariance matrices. As shown in Fig. \ref{simmap}, our method can get more reasonable similarity maps than cosine similarity. This also explains why DACM can converge faster than existing methods. Furthermore, we also visualize the learned deformable offsets for the support and query dimensions in Fig. \ref{vis_defor}. It can be seen that the deformable attention mechanism tends to attend on informative regions to learn more powerful representations.

\section{Conclusions}
In this paper, we present a few-shot segmentation method by doubly deformable aggregating covariance matrices. The proposed method aggregates learnable covariance matrices with a doubly deformable 4D Transformer to predict segmentation results effectively. Specifically, we make the following contributions. 1) We devise a novel hard example mining mechanism for learning covariance kernels of the Gaussian process to enable a more accurate correspondence measurement. 2) We design a doubly deformable 4D Transformer to effectively and efficiently aggregate the multi-scale cost volume into the final segmentation results. 3) By combining these two modules, the proposed method can achieve state-of-the-art few-shot segmentation performance with a fast convergence speed.

\section{Acknowledgement}
This work is jointly supported by the European Research Council (ERC) under the European Union's Horizon 2020 research and innovation program (grant agreement No. [ERC-2016-StG-714087], Acronym: \textit{So2Sat}), by the Helmholtz Association
through the Framework of Helmholtz AI (grant  number:  ZT-I-PF-5-01) - Local Unit ``Munich Unit @Aeronautics, Space and Transport (MASTr)'' and Helmholtz Excellent Professorship ``Data Science in Earth Observation - Big Data Fusion for Urban Research''(grant number: W2-W3-100), by the German Federal Ministry of Education and Research (BMBF) in the framework of the international future AI lab "AI4EO -- Artificial Intelligence for Earth Observation: Reasoning, Uncertainties, Ethics and Beyond" (grant number: 01DD20001) and by German Federal Ministry of Economics and Technology in the framework of the "national center of excellence ML4Earth" (grant number: 50EE2201C).
\clearpage

\title{Supplementary Material}
\titlerunning{DACM for few-shot segmentation}
\author{Zhitong Xiong \orcidlink{0000-0002-3953-585X} \inst{1} \and
Haopeng Li \orcidlink{0000-0001-8175-5381} \inst{2} \and
Xiao Xiang Zhu \orcidlink{0000-0001-5530-3613} \inst{1,3}}
\authorrunning{Z. Xiong et al.}
\institute{Data Science in Earth Observation, Technical University of Munich (TUM)\\ \and
School of Computing and Information Systems, University of Melbourne \and
Remote Sensing Technology Institute (IMF), German Aerospace Center (DLR)\\}
\maketitle
\begin{figure} [!]
	\centering
	\includegraphics[width=0.88\columnwidth]{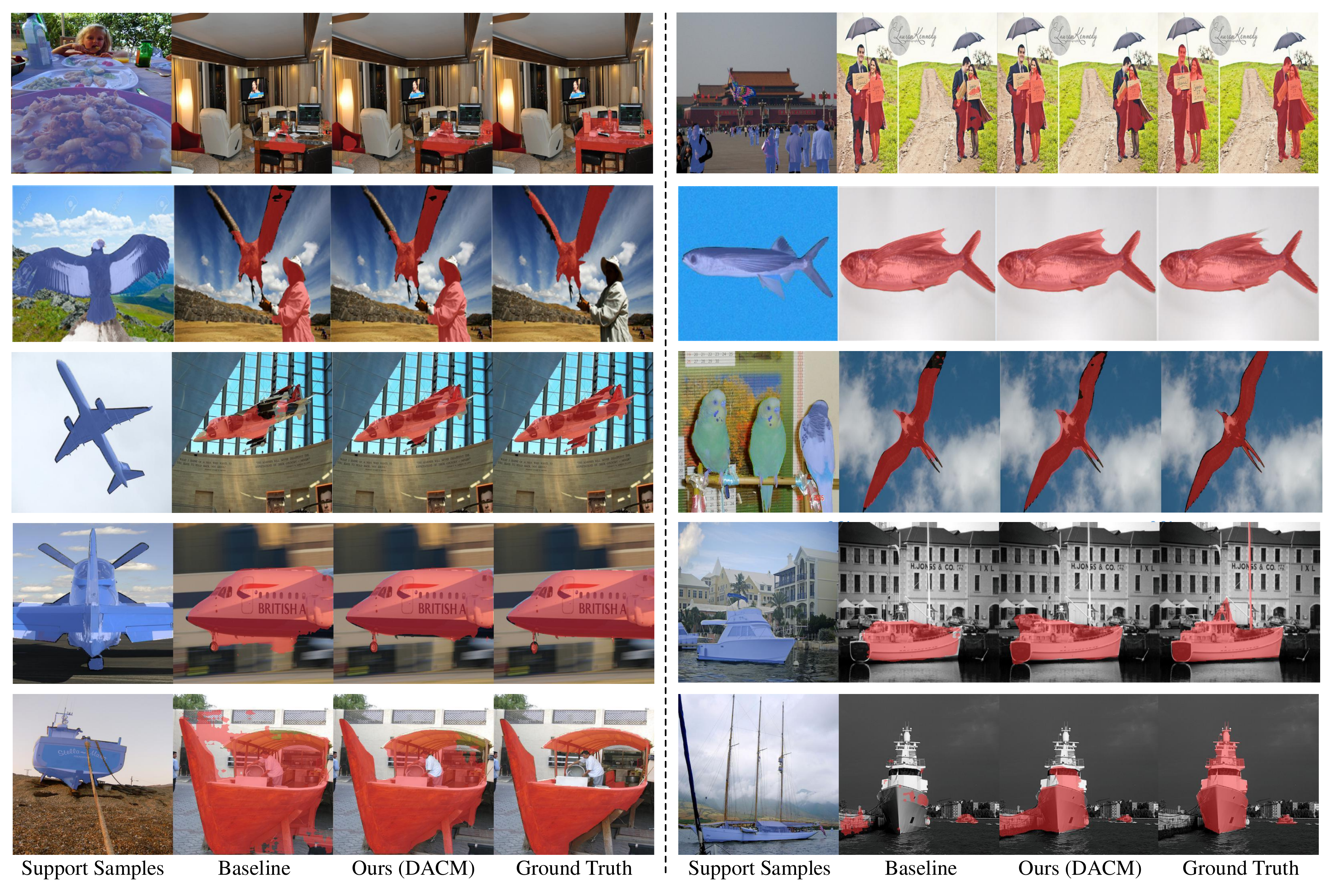}
	\caption{Visualization of the few-shot segmentation results on PASCAL-$5^i$, COCO-$20^i$ \cite{lin2014microsoft}, and FSS-1000 \cite{li2020fss} datasets. From left to right: support samples, results of the baseline method, results of the DACM, and ground truth labels. It can be clearly seen that our method can output better results than the baseline method. 
}\label{GP3}
\end{figure}
\section{Few-shot Segmentation Visualization}
Three datasets for few-shot segmentation are exploited to evaluate the proposed method, including the PASCAL-$5^i$ \cite{shaban2017one}, COCO-$20^i$ \cite{lin2014microsoft}, and FSS-1000 \cite{li2020fss}. 
As we can see in Fig. \ref{GP3}, there exist large variations between the support images and the query images in scale, pose and appearance.

\begin{figure*}[t]
	\centering
	\includegraphics[width=0.96\textwidth]{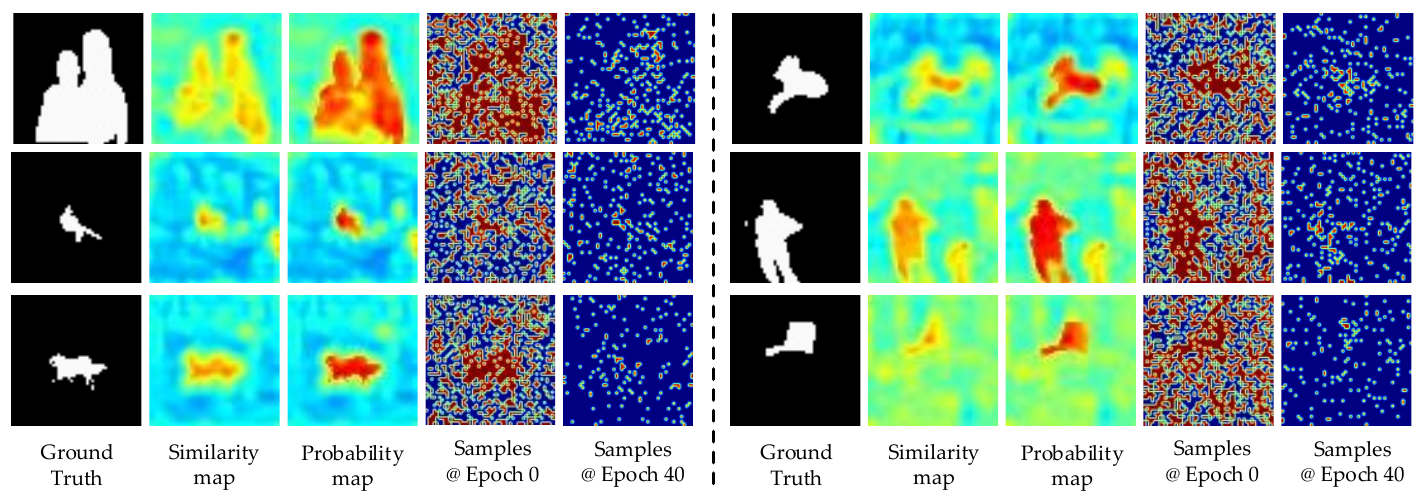}
	\caption{Visualization of some qualitative examples of the proposed hard example-aware sampling strategy for training GP models of the DACM model.
}
	\label{sample_more}
\end{figure*} 
% the scale, pose and appearance variations between support images and query images are large. 
Although only 50 epochs are used for training the proposed DACM method, the performance of our model is clearly better than that of the baseline method for some difficult test samples. Compared with HSNet \cite{min2021hypercorrelation}, qualitative results in Fig. \ref{GP3} show that DACM can obtain more complete segmentation maps owing to larger receptive fields of the Transformer architecture. In Fig. \ref{sample_more}, we also visualize more examples of sampled locations during the training process of our DACM method.

\section{Covariance Matrices Analysis}

\begin{figure}
	\centering
	\includegraphics[width=0.8\columnwidth]{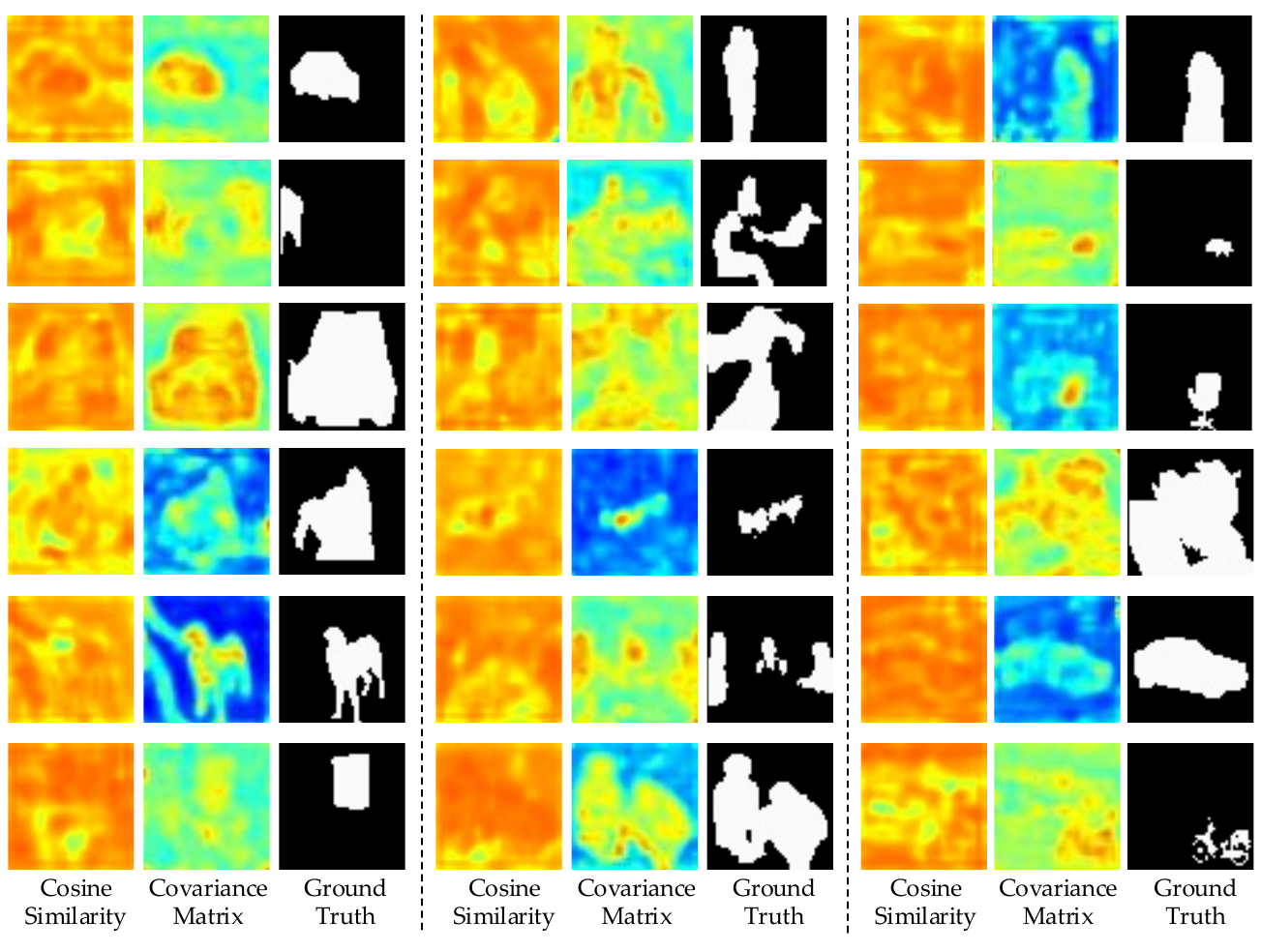}
	\caption{Visualization of the learned variance matrices. For a clear visualization, we pool the matrices into 2D similarity map along the support dimension. It can be seen that similarity maps of DACM are more consistent with the ground truth label.}
	\label{more_sim}
\end{figure}

Different from HSNet \cite{min2021hypercorrelation}, we use the learned covariance kernel functions to measure the similarity between the support features and the query features instead of a fixed cosine similarity. The fixed cosine similarity is computed as follows:
\begin{equation}
    \bm{C}_1(\bm{x}_q,\bm{z}_s) = \text{ReLU}\left(\frac{\bm{x}_q^\mathrm{T}\bm{z}_s}{||\bm{x}_q||||\bm{z}_s||}\right),
\end{equation}
where $\bm{x}_q$ is the query feature, and $\bm{z}_s$ is the masked support feature.

Theoretically, the cosine similarity used in \cite{min2021hypercorrelation} can be viewed as a special case of our method, \emph{i.e.}, using a fixed linear kernel. For a more effective similarity measurement, we target at learning the kernel functions $k(\cdot,\cdot)$ on different datasets using Gaussian process. The covariance matrices can be computed as follows:
\begin{equation}
    \bm{C}_2(\bm{x}_q,\bm{z}_s) = \text{ReLU}\left(k\left(\frac{\bm{x}_q}{||\bm{x}_q||},\frac{ \bm{z}_s}{||\bm{z}_s||}\right)\right).
\end{equation}
To better understand the learned kernel functions in GP, we have visualized the covariance matrices in Fig. \ref{more_sim}. For the cosine similarity map $\bm{C}_1 \in \mathbb{R}^{H_q\times W_q\times H_s\times W_s}$, we reshape it into $\bm{C}_1 \in \mathbb{R}^{H_q \times W_q\times H_s W_s}$ and sum it up along the third dimension. Then we can get 2D similarity maps of the query image. The 2D similarity maps are shown in Fig. \ref{more_sim}. For the covariance matrices $\bm{C}_2 \in \mathbb{R}^{H_q\times W_q\times H_s\times W_s}$, we conduct the similar computation as the cosine similarity map $\bm{C}_1$. 

From Fig. \ref{more_sim} we can see that covariance matrices learned by the DACM model are more reasonable than the simple cosine similarity map. Then, the learned similarity maps can be processed by the proposed DDT model to further enhance the performance of few-shot segmentation.

\begin{figure*}[t]
	\centering
	\includegraphics[width=0.6\textwidth]{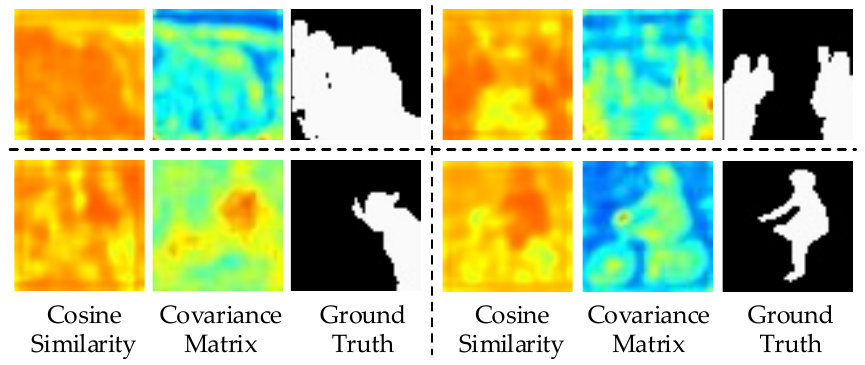}
	\caption{Visualization of some failure cases of the learned covariance matrices. Compared with the cosine similarity, it can be  seen that the learned covariance matrices fail to separate the foreground objects and the background.
}
	\label{fail_sim}
\end{figure*} 
\section{Limitations and Failure Cases}
In this subsection, we will visualize and analyze some failure cases of the proposed DACM method. In Fig. \ref{fail_sim}, we can see some failure cases of the learned covariance matrices. It can be seen that, the similarity between the support and the background of query samples is clearly low. However, in some cases, the similarity between the support and foreground objects of query samples is incorrectly high. From the visualization results, we can find that although the global representation learning of Transformer is beneficial in most cases, there still exist some situations where smaller receptive fields work better. 

\begin{figure*}[t]
	\centering
	\includegraphics[width=0.9\textwidth]{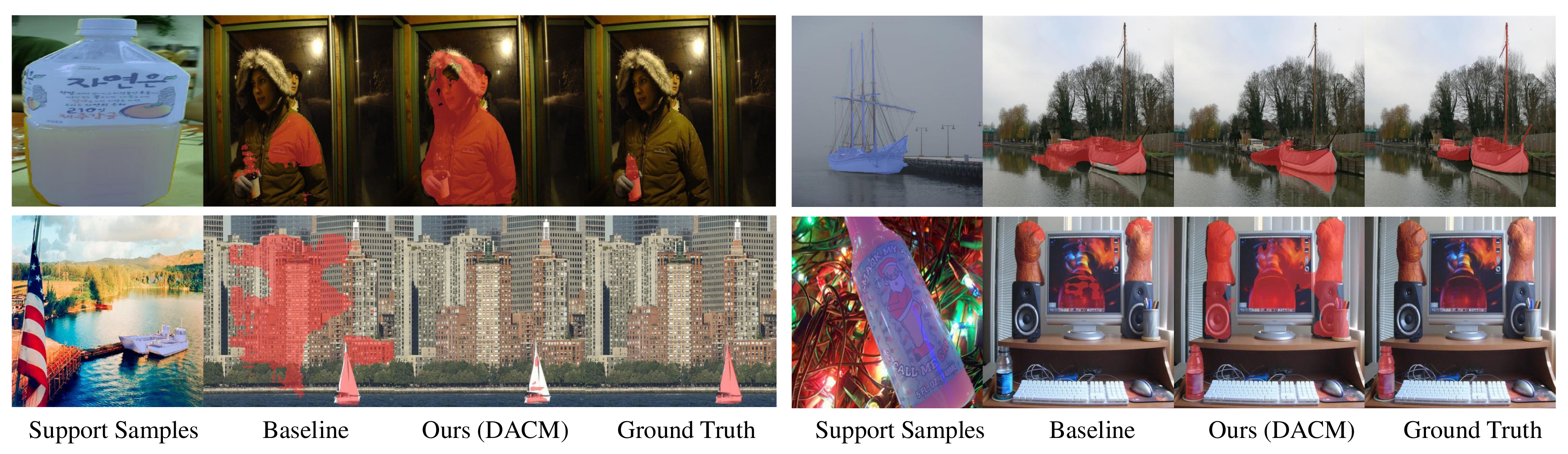}
	\caption{Visualization of some failure cases of 1-shot segmentation results.
}
	\label{failcase}
\end{figure*}

In Fig. \ref{failcase}, we also visualize some segmentation failures. As for the top-left image, the man holding the bottle is misclassified as a foreground object. Take the top-right result as an example, DACM incorrectly classify the reflection of the boat as a foreground object. Note that the baseline method obtains better mIoU results on these visualized samples than DACM. In the future work, we believe that studying how to effectively combine global and local representations can further enhance the performance of the few-shot segmentation task.

\section{Details of Gaussian Process}

Gaussian process (GP) is a non-linear and non-parametric Bayesian model for regression and classification \cite{williams1996gaussian}. By exploiting the correlation between data samples, Gaussian process performs probabilistic non-linear prediction that is described by Gaussian distribution with estimated mean and covariance. We explain the details of general GP as follows. Formally, the data set consists of $N$ samples of dimension $D$, \textit{i.e.}, $\{X=\{{x}_i\}_{i=1}^N,\bm{y}=\{y_i\}_{i=1}^N\rbrace$, where ${x}_i\in \mathbb{R}^D$ is a data point and $y_i$ is the corresponding label. The GP regression model assumes that the outputs $y_i$ can be regarded as certain deterministic latent function $f({x}_i)$ with zero-mean Gaussian noise $\varepsilon$, \textit{i.e.}, $y_i = f({x}_i)+\varepsilon$, where $\varepsilon\sim\mathcal{N}(0,\sigma^2)$. GP sets a zero-mean prior on $f$, with covariance $k({x}_i,{x}_j)$. The covariance function (kernel) $k$ reflects the smoothness of $f$. Commonly-used kernels can be found in Section \ref{123}. The hyper-parameters in the kernels are optimized by maximizing the marginal likelihood of the training data, which is given as follows,
\begin{equation}
p(\bm{y}| X)=\int p(\bm{y} | \bm{f}, X) p(\bm{f} | X) \mathrm{d} \bm{f}, \\
\end{equation}
where $\bm{f}=[f(x_1),f(x_2),\cdots,f(x_N)]^\mathrm{T}$.
Here, the term marginal likelihood refers to the marginalization over the function values $\bm{f}$.
Setting the prior of the Gaussian process model $p(\bm{f}| X)$ to be a Gaussian distribution $\mathcal{N}(\bm{0},\bm{K})$,
the marginal log-likelihood $\log p(\bm{y}|X)$ can be expressed by,
\begin{equation}
\log p(\bm{y} | X) = -\frac{1}{2}\log| \bm{K}|-\frac{1}{2}\bm{y}^\mathrm{T}(\bm{K}+\sigma^2\bm{I}_N)^{-1}\bm{y}-\frac{N}{2}\log 2\pi, \label{e9}
\end{equation}
where $\bm{y}=[y_1,y_2,\cdots,y_N]^\mathrm{T}$, $\bm{K}_{ij}=k({x}_i,{x}_j)$, and $\bm{I}_N$ is the identity matrix of size $N$. For example, the Automatic Relevance Determination Squared Exponential (ARD SE) kernel function is defined as
\begin{equation}
k({x}_i,{x}_j)=\sigma_0^2\exp\left\lbrace -\frac{1}{2}\sum_{d=1}^D\frac{(({x}_i)_d-({x}_j)_d)^2}{l_d^2}\right\rbrace,
\label{eq}
\end{equation}
where $\sigma^2, \sigma_0^2,\left\lbrace l_d\right\rbrace_{d=1}^D$ are hyper-parameters. By maximizing Eq. \ref{e9}, the hyper-parameters in Eq. \ref{eq} for computing the covariance matrices can be optimized.

After the hyper-parameters are optimized, the predictive distribution for the test case $x^*$ can be calculated in a closed-form as follows:
\begin{align}
    &\mu_{y^*}=\bm{k}^\mathrm{T}(\bm{K}+\sigma^2\bm{I}_N)^{-1}\bm{y},\\
    &\sigma_{y^*}^2 = {k}^*-\bm{k}^\mathrm{T}(\bm{K}+\sigma^2\bm{I}_N)^{-1}\bm{k}+\sigma^2,
\end{align}
where $\bm{k}=[k(x_1,x^*),k(x_2,x^*),\cdots,k(x_N,x^*)]^\mathrm{T}$ and $k^*=k(x^*,x^*)$.

\begin{algorithm}[h]
\SetAlgoLined
    \PyComment{Gaussian process implementation details} \\
    \PyCode{class ExactGPModel(models.ExactGP):} \\
    \Indp
        \PyCode{def init(self, kernel='rbf'):} \\
        \Indp
            \PyCode{self.mean\_module = means.ConstantMean()} \PyComment{using constant mean} \\
            \PyComment{Different Kernels} \\
            \PyCode{if kernel == 'linear':} \\
            \Indp
                \PyComment{Linear kernel is equivalent to cosine similarity.}\\
                \PyCode{self.linear\_kernel = kernels.LinearKernel()} \\
            \Indm
            \PyCode{if kernel == 'rbf':} \\
            \Indp
                \PyComment{RBF kernel with automatic relevance determination.}\\
                \PyCode{self.rbf\_kernel = kernels.RBFKernel(ARD)} \\
            \Indm
            \PyCode{if kernel == 'additive':} \\
            \Indp
                \PyComment{Additive kernel by summing over multiple kernels.}\\
                \PyCode{self.covar\_module = kernels.LinearKernel() + kernels.RBFKernel(ARD)} \\
            \Indm
        \Indm
    \Indm
    
    \Indp
        \PyCode{def forward(self, x):} \\
        \Indp
            \PyComment{Mean function computation} \\
            \PyCode{mean\_x  = self.mean\_module(x)} \\
            \PyComment{Covariance function computation} \\
            \PyCode{covar\_x = self.covar\_module(x)} \\
            \PyCode{return distributions.MultivariateNormal(mean\_x, covar\_x)} \\
        \Indm
    \Indm
\caption{Pseudocode for implementation of the Gaussian process}
\label{algo:your-algo}
\end{algorithm}

\section{Gaussian Process Implementation}
\label{123}
In this section, we present the Pytorch-style implementation of the GP model. The most critical part of the GP model is the choice of covariance kernel functions. In algorithm \ref{algo:your-algo}, we show three types of kernel functions, including the linear kernel, the radial basis function (RBF) kernel and the additive mixed kernel. 

For two vectors $\mathbf{x}_{\mathbf{1}},\mathbf{x}_{\mathbf{2}}$ in the feature map, the linear kernel, the RBF kernel, and the additive mixed kernel, with hyperparameters $\{v, \Theta\}$, are defined as follows:
\begin{equation}
\begin{aligned}
    &k_{\text {Linear }}\left(\mathbf{x}_{\mathbf{1}}, \mathbf{x}_{\mathbf{2}}\right)=v \mathbf{x}_{\mathbf{1}}^{\top} \mathbf{x}_{\mathbf{2}}, \\
    &k_{\mathrm{RBF}}\left(\mathbf{x}_{\mathbf{1}}, \mathbf{x}_{\mathbf{2}}\right)=\exp \left(-\frac{1}{2}\left(\mathbf{x}_{\mathbf{1}}-\mathbf{x}_{\mathbf{2}}\right)^{\top} \Theta^{-2}\left(\mathbf{x}_{\mathbf{1}}-\mathbf{x}_{\mathbf{2}}\right)\right),\\
    &k_{\text {Additive }}\left(\mathbf{x}_{\mathbf{1}}, \mathbf{x}_{\mathbf{2}}\right)=k_{\text {Linear }}\left(\mathbf{x}_{\mathbf{1}}, \mathbf{x}_{\mathbf{2}}\right) + k_{\mathrm{RBF}}\left(\mathbf{x}_{\mathbf{1}}, \mathbf{x}_{\mathbf{2}}\right).
\end{aligned}
\end{equation}
%.
Although the RBF kernel is used in this work, here we show that the proposed framework can be easily extended by using different kernel types or their combinations.

\bibliographystyle{splncs04}
\bibliography{egbib}
\end{document}